\documentclass[twoside,11pt]{article}

\usepackage{blindtext}

\usepackage{siunitx}
\usepackage{amsmath}
\usepackage{bbm}
\usepackage{mathtools}
\usepackage{booktabs}
\usepackage{multirow}
\usepackage{xfrac}
\usepackage{xstring}
\usepackage{xstring}
\usepackage{xpatch}
\usepackage{jmlr2e}
\usepackage[section]{placeins}

\newcommand{\R}{\mathbb{R}}

\usepackage{multirow}
\usepackage{makecell}
\usepackage{array}

\DeclareMathOperator{\argmin}{arg\,min}
\DeclareMathOperator{\argmax}{arg\,max}
\let\mid\vert
\NewExpandableDocumentCommand\mcc{m}{\multicolumn{1}{c}{#1}}

\newcommand{\p}{\Delta p_{t,\textrm{rel}}}

\usepackage{lastpage}
\usepackage{amsmath}
\usepackage{amssymb}

\ShortHeadings{How Well do Generative Models Solve Inverse Problems?}{Krüger, Materne, Krebs and Gottschalk}
\firstpageno{1}
\begin{document}

\title{How well do generative models solve inverse problems?\\ -- A benchmark study --}

\author{\name Patrick Krüger${}^1$ \email krueger@math.tu-berlin.de \\
\name Patrick Materne${}^1$ \email patrick.materne@campus.tu-berlin.de\\
\name Werner Krebs${}^2$ \email wernerkrebs@siemens-energy.com\\
\name Hanno Gottschalk${}^1$ \email gottschalk@math.tu-berlin.de\\
\vspace{.3cm}\\
       \addr ${}^1$Institute of Mathematics,
       TU Berlin, Straße des 17.\ Juni 136,
       10623 Berlin, Germany\\
       ${}^2$Siemens Energy, Rheinstr.\ 100,
       45478 Mülheim, Germany
}

\maketitle
\thispagestyle{empty}
\begin{abstract}
Generative learning generates high dimensional data based on low dimensional conditions, also called prompts. Therefore, generative learning algorithms are eligible for solving (Bayesian) inverse problems. In this article we compare a traditional Bayesian inverse approach based on a forward regression model and a prior sampled with the Markov Chain Monte Carlo method with three state of the art generative learning models, namely conditional Generative Adversarial Networks, Invertible Neural Networks and Conditional Flow Matching. We apply them to a problem of gas turbine combustor design where we map six independent design parameters to three performance labels. We propose several metrics for the evaluation of this inverse design approaches and measure the accuracy of the labels of the generated designs along with the diversity. We also study the performance as a function of the training dataset size. Our benchmark has a clear winner, as Conditional Flow Matching consistently outperforms all competing approaches. 
\end{abstract}


\section{Introduction}
Engineering may be understood as the discipline of designing devices that satisfy prescribed technical requirements. Once the design space $\mathcal{X}$ has been defined, the central task is to identify a parameter vector $x$ such that an evaluation function $s(x)$, which produces a vector of performance labels $y$ in label space $\mathcal{Y}$, yields values sufficiently close to the target specifications.

In many practical applications, the evaluation function involves computationally expensive numerical simulations, such as finite element analysis (FEA) or computational fluid dynamics (CFD), which model the physical state of a component. These simulations are typically followed by a post-processing step that extracts the relevant performance metrics. From this perspective, the engineering design problem can be formulated as a (Bayesian) inverse problem: Given a desired performance vector $y$, the goal is to determine its pre-image $\{x\in\mathcal{X} : s(x) = y\}$ under the evaluation function, using as few observations as possible.

These observations are often noisy and can be expressed as
\begin{equation}
\label{eqa:regression}
Y^{(j)} = s(X^{(j)}) + \varepsilon_j, \quad i = 1, \ldots, n,
\end{equation}
where $\varepsilon_j$ denotes a noise term accounting for measurement or simulation uncertainty. Here we follow the convention from probability theory that random variables like the (noisy) label $Y$ or the (sampled) set of design parameters are written in capital letters $X,Y$, whereas concrete realizations $x\in\mathcal{X}$, $y\in\mathcal{Y}$ are written in lower case. Together, the function $f$ and the statistical properties of the noise $\varepsilon_i$ define a conditional probability distribution $p(y \vert x)$ of the performance labels given the design parameters. For instance, assuming that the noise is normally distributed with variance $\sigma^2 > 0$ and identity covariance matrix $\mathbbm{1}$ in label space, i.e., $\varepsilon_j \sim \mathcal{N}(0, \sigma^2 \mathbbm{1})$, the resulting conditional distribution is
\[
p(y \vert x) = \mathcal{N}(s(x), \sigma^2 \mathbbm{1}).
\]

Machine learning models have been used for a long time to assist engineering tasks. Surrogate modeling enables faster design space exploration by substituting costly function evaluations $s(x)$ with inexpensive evaluations of a learned function $f_\theta(x)$ that has been trained on realizations $(x^{(j)},y^{(j)})$. This leads to a learned representation $p_\theta(y|x)$ with $p_\theta(y|x)=N(f_\theta(x),\hat\sigma^2\mathbbm{1})$ where $\hat\sigma^2$ is a standard variance estimator. It is well known that this model, when trained with the negative log-likelihood as loss function, amounts to standard least squares regression.
The conditional distribution $p_\theta(y|x)$ can now be used to sample $Y$ for a large set of realizations $x$ of $X$. The hope is to find some $x\in\mathcal{X}$ that have a corresponding label $y\approx y^*$.  This amounts to a brute force forward search. More sophisticated methods rely on the maximum a posteriori (MAP) principle to infer the most probable label $\hat{y}(x)$, followed by an optimization step that determines the corresponding design variable by minimizing the discrepancy between the predicted label and the desired target value:\begin{equation}
\label{eqa:invProblem}
\hat{y}(x) = \arg\max_{y\in \mathcal{Y}} p_\theta(y \mid x),
\qquad \text{ and }
\hat{x} \in \arg\min_{x\in\mathcal{X}} \| \hat{y}(x) - y^* \|^2 .
\end{equation}
Here the solution strategy can be built on local or global optimization per choice of the designer, e.g. when $p_{\theta}(y|x)=N(f_\theta(x),\hat\sigma^2\mathbbm{1})$, $\hat y(x)=f_\theta(x)$ solves the first problem and the second problem amounts to finding  $\hat x\in \argmin_{x\in\mathcal{X}}\|\hat f_\theta(x)-y^*\|^2$ via optimization. 

Inverse problems like \autoref{eqa:invProblem} and their regularization against ill-posedness have been widely studied \cite{tikhonov1995numerical}.  Also, refined probabilistic modeling strategies for goal oriented searches, e.g.\ based on Gaussian processes, have been proposed \cite{forrester2008engineering}. This approach however has  two disadvantages: First, it requires an involved optimization that has to be skillfully set up. Second, each optimization run yields only a single proposed design, and exploration of the set \(\{x\in\mathcal{X} : s(x) \approx y^*\}\) is, in general, not exhaustive.

One major solution approach to inverse problems is to apply the Bayesian formula to the conditional probability distribution $p_\theta(Y|X)$, see e.g.\ \cite{tarantola2005inverse}. By specification of an prior probability density $p_X(x)$ on the design space, we obtain the a posteriori probability for $x$:
\begin{equation}
    \label{eqa:a_posteriori_probability}
    p_\theta (x|y)=p_\theta(y|x) \,\frac{p_X(x)}{p_Y(y)}.
\end{equation}
Here $p_Y(y)$ is the marginal probability density over the label space $\mathcal{Y}$. While the computation of $p_Y(y)$ is oftentimes intractable, it is needed neither for finding design proposals $\hat x\in \argmax _{x\in\mathcal{X}} p_\theta(x|y^*)$, nor for sampling from the posterior distribution with Markov Chain Monte Carlo (MCMC) methods \cite{gamerman2006markov}, which in principle will enable an extensive retrieval of valid design proposals. However, MCMC methods are not trivial to set up and often require expert surveillance, making full automation difficult.\\
\ \\
This naturally raises the question of why one would not simply exchange the roles of \(X\) and \(Y\) in \autoref{eqa:regression} and learn \(p( x\mid y)\) directly from the data pairs \((x^{(j)}, y^{(j)})_{j=1}^n\). For a long time, this approach was hindered by model misspecification. While the normal distribution is suitable as a model for the error distribution in regression with low dimensional output, it is clearly unsuitable to model a distribution in design space that is concentrated around the highly non-linear and potentially multimodal subset $\{ x\in\mathcal{X}: s(x)=y^*\}$. However, learning high-dimensional distributions conditioned on lower-dimensional input, together with efficient sampling from the resulting conditional distributions, lies squarely within the scope of modern generative learning. Numerous generative modeling approaches have been proposed, including energy-based methods, Variational Autoencoders \cite{kingma2013auto}, Generative Adversarial Networks \cite{goodfellow2014gans} and diffusion-based models, as well as flow-based models such as  Normalizing Flows \cite{rezende2015variational,chan2023lu} and Conditional Flow Matching (CFM) \cite{lipman2022flow}. Invertible Neural Networks \cite{ardizzone2018analyzing} demonstrated the use of (discrete-time) normalizing flows to solve inverse problems.

As discussed above, all of these methods are, in principle, capable of learning solutions to the Bayesian inverse problem. The remaining questions concern which approach does so most efficiently, whether any of the modern generative methods outperform the traditional Bayesian inverse approach based on \autoref{eqa:regression} and \autoref{eqa:a_posteriori_probability}, and how much data is required to train the respective models effectively.\\
In this paper, we study this set of questions and compare the Bayesian inverse approach with conditional Wasserstein-GAN, INN and CFM. We apply these methods to an inverse design task of a gas turbine combustor and develop metrics that not only account for the precision of generated designs, but also for the diversity of designs, i.e. the coverage of the space of design alternatives $\{x\in\mathcal{X}:s(x)=y^*\}$ by the retrieved design proposals. We find that we have a clear winner of the benchmark study, as CFM outperforms all other models in terms of data efficiency and precision while maintaining diversity of retrieved designs.

Our paper is organized as follows: In \autoref{sec:related_work} we shortly review the development and state of the art of generative learning. \autoref{sec:dataset} introduces our inverse design problem following \cite{krueger2025generative}, as well as the resulting training dataset. \autoref{sec:Theory} introduces the models we consider in this benchmark study. \autoref{sec:implementation} explains our training and validation procedures, whereas \autoref{sec:Validation} presents our evaluation approach and numerical results. We conclude and give an outlook on future research in \autoref{sec:conclusion}.

\section{Related Work}
\label{sec:related_work}
A wide range of machine learning approaches, including those considered in this work, have been applied to inverse problems across various application domains. Since their introduction, Normalizing Flows (NF, \cite{kingma2018glow}) have been extensively employed in conditional image generation (\cite{zhai2024normalizing}, \cite{xiao2019method}), as well as in related tasks such as colorization \cite{10448464}, inpainting \cite{10.1007/978-3-031-20050-2_4}, and image enhancement \cite{Wang_Wan_Yang_Li_Chau_Kot_2022}. Within the engineering domain, NFs have also been used for inverse design problems, for example in the design of thermophotovoltaic emitters \cite{yang2023normalizing}.

Building on the same architectural principles, Invertible Neural Networks (INN, \cite{ardizzone2018analyzing}) further specialize the concept of normalizing flows with inverse problems as a primary focus. In addition to applications in conditional image generation (\cite{ardizzone2019guided}, \cite{10.1007/978-3-030-71278-5_27}), INNs have been successfully applied to engineering design tasks, including the inverse design of gas turbine combustors \cite{krueger2025generative}, compressor fan blades \cite{jia2025inverse}, and airfoils \cite{glaws2022invertible}.

Flow-based models were further extended through the introduction of time-continuous formulations such as neural ordinary differential equations (NODE, \cite{chen2018neural}). In this context, conditional flow matching (CFM, \cite{lipman2022flow}, \cite{tong2023conditional}) provides a scalable training framework for conditional generative modeling. Despite the relative novelty of this approach, CFM has already been applied to a variety of inverse design problems, including the design of RNA sequences \cite{nori2024rnaflow}, peptides \cite{li2024full}, and ship propellers \textbf{ref}.

Generative adversarial networks (GAN, \cite{goodfellow2020generative}) constitute a prominent alternative to flow-based models in generative learning. Although GANs are not inherently designed for inverse problems, conditional variants such as conditional (Wasserstein) GANs (\cite{samrat2020conditional}, \cite{arjovsky2017wassersteingan}) have been successfully applied to inverse modeling tasks, including digital rock reconstruction \cite{zheng2022digital}, cellular structure generation \cite{WANG2022115060}, and turbulence modeling \cite{Drygala_2022}.

Beyond the methods discussed above, several additional machine learning techniques have been explored for inverse problems. Variational autoencoders (VAE, \cite{kingma2013auto}) learn compressed latent representations of complex data and have been applied, for example, to photoacoustic tomography \cite{doi:10.1137/22M1489897} and wind turbine airfoil design \cite{yang2023inverse}. Other examples include convolutional neural networks \cite{sekar2019inverse}, physics-informed neural networks \cite{doi:10.1137/21M1397908}, and direct forward and inverse modeling using multilayer perceptrons \cite{zhou2022design}. A comprehensive review covering these approaches, as well as active learning, Bayesian optimization, and Gaussian processes in the context of materials design, is provided by \cite{lee2023machine}.

\section{Inverse Problem Description/ The Combustor Dataset}\label{sec:dataset}

The dataset utilized in this work was initially developed for inverse \(\mathrm{H}_2\)-combustor design in the field of mechanical engineering, as described in \cite{krueger2025generative}. Each data point $(X, Y)$ corresponds to a parameterized combustor geometry that is uniquely defined by a vector $X$ of independent design parameters, while $Y$ is a vector of associated performance metrics obtained via computational fluid dynamics (CFD) simulations on the geommetry defined by $X$. Each combustor geometry consists of a combustion chamber plenum, a premix tube, and the combustion chamber itself. Fuel is introduced through a central lance extending from the plenum into the premix tube. The premix tube includes multiple injection ports and vortex generators to enhance fuel-air mixing. \autoref{fig:Figure_Geometry} illustrates the overall geometry.\\

To facilitate a comprehensive exploration of the design space and ensure efficient expert validation, a reduced number of six independent design parameters (\autoref{eq:deisgn_parameters}) selected based on domain expertise: 

\begin{figure}[htbp]
    \centering
    \includegraphics[width=1\textwidth]{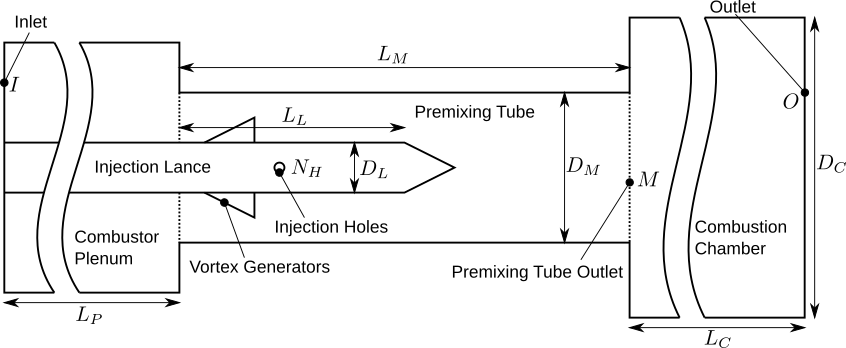}
    \caption{Schematic of the parameterized combustor geometry defined by independent variables $X$. Source: \cite{krueger2025generative}}
    \label{fig:Figure_Geometry}
\end{figure}

\begin{equation}
\label{eq:deisgn_parameters}
    X = (X_i)_{i\in\{1,\cdots,6\}}= (R_A, N_H, D_M, R_D, R_L, L_P).
\end{equation}

Here, \(R_A\), \(N_H\), and \(D_M\) define the vortex generator area ratio, the number of injection ports, and the premix tube diameter, respectively. The parameter \(R_D\) denotes the ratio of the fuel lance diameter \(D_L\) to the premix tube diameter \(D_M\), while \(R_L\) represents the length-to-diameter ratio of the premix tube, and \(L_P\) specifies the plenum length. An overview of independent parameters and their respective ranges is given by \autoref{t_indep_params}. \\

\begin{table}
\centering
    \begin{tabular}{m{0.1\textwidth}m{0.5\textwidth}m{0.2\textwidth}}\label{t_indep_params}
         \textbf{Parameter}&\mcc{\textbf{Definition}}&\textbf{Range}\\
         
         \hline
         $R_A$ & Ratio of free area at vortex generators to cross sectional area of premixing tube&$[0.63,0.83]$ \\
         \hline
         $N_H$ & Number of fuel injection holes on injection lance&$[2,10]$ \\
         \hline
         $D_M$& Diameter of premixing tube in \qty{}{mm}&$[20,45]$ \\
         \hline
        $R_D$ & Ratio of diameter of lance to diameter of premixing tube&$[0.35,0.55]$ \\
         \hline
         $R_L$ & Ratio of length of premixing tube to diameter of premixing tube&$[4,12]$ \\
         \hline
         $L_P$ & Length of combustor plenum in \qty{}{mm}&$[200,900]$\\
    \end{tabular} 
    \caption{Independent parameters $X_i$ of the Combustor dataset.}
\end{table}
All remaining geometrical characteristics are derived from these six parameters. For instance, the premix tube length \(L_M\) is computed as $L_M = D_M \cdot R_L$, while length and diameter of the fuel lance are given by $D_L = R_D \cdot D_M$ and $L_L = 0.2 \cdot L_M$, respectively. 
The combustion chamber length is fixed at \(L_C = 1000\,\mathrm{mm}\) with a discharge ratio of \(R_C = 4\). The number of vortex generators are placed between injection holes, thus their number is $N_V=\min(\frac{N_H}{2},2)$.\\

\begin{figure}[htbp]
    \centering
    \includegraphics[width=1\textwidth]{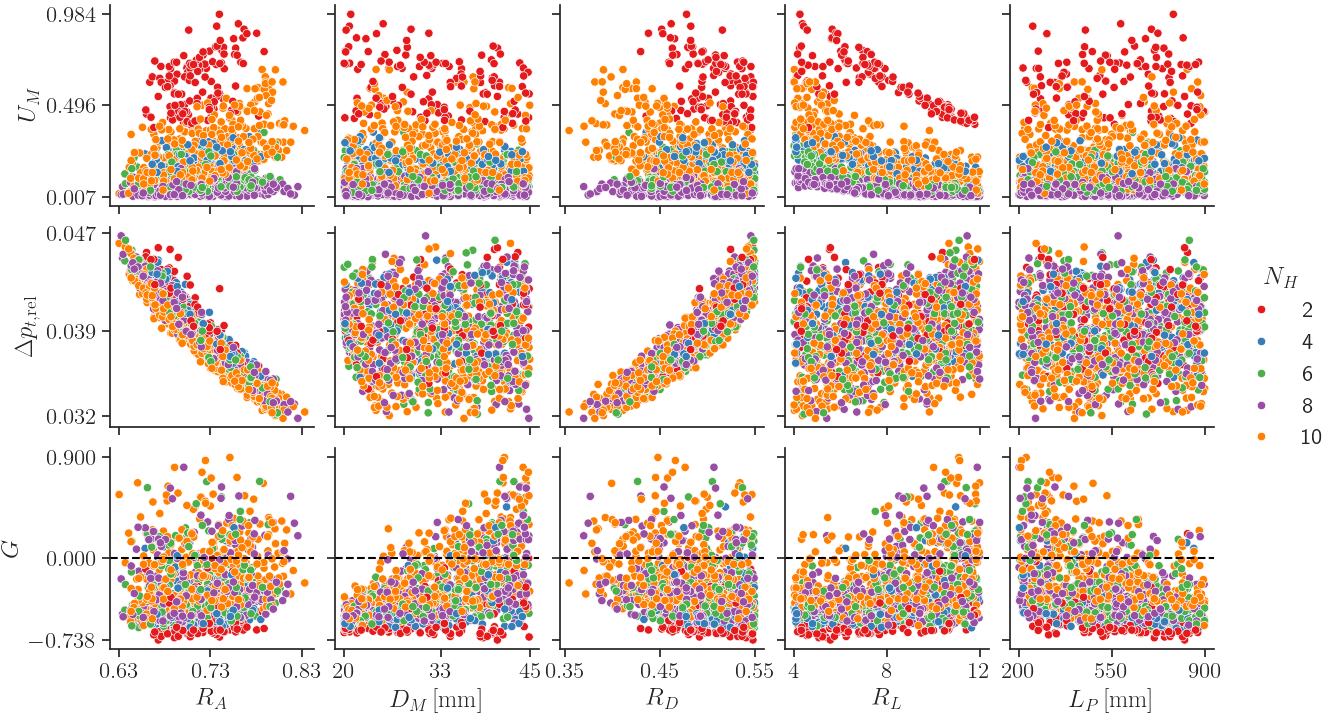}
    \caption{Scatter plot of the combustor dataset \(\mathcal{D}\), showing relationships between design parameters $X_i$ and performance metrics $Y_i$. Dashed line: stability boundary \(G = 0\). Source: \cite{krueger2025generative}.}
    \label{fig:Figure_Dataset}
\end{figure}

Performance of geometries defined by parameters $X$ is characterized by a vector $Y$ sonsisting of three performance labels:
\begin{equation}\label{eq:labels}
    Y = (Y_i)_{i\in\{1,2,3\}}=(U_M, \Delta_{pt,re}, G).
\end{equation}
in \autoref{eq:labels}, the unmixedness \(U_M\), measures the degree of fuel-air inhomogeneity at the premix tube outlet, which is a key indicator of expected \(\mathrm{NO}_x\) emissions. The relative pressure drop $\p$ betweem system inlet and outlet reflects the efficiency of the system and its compatibility with cooling technologies. Finally, the thermoacoustic growth rate \(G\) characterizes the stability of the combustion process—positive values suggest potential instability due to self-amplifying acoustic feedback, whereas negative values indicate a stable regime. $U_M$ and $\p$ are derived from steady-state, reacting, turbulent RANS simulations using Siemens Simcenter STAR-CCM+ \cite{siemens2019star}. The simulations employ the \(k\)-\(\omega\)-SST turbulence model and a Flamelet Generated Manifold (FGM) approach based on GRI3.0 chemistry for \(\mathrm{CH}_4\)/air combustion. $G$ is evaluated using the GeneAC  network solver \cite{krebs2013comparison}, which models acoustic wave propagation via linearized Euler equations \cite{kru2001prediction}.\\
\ \\
To ensure an even distribution across the domain, parameter vectors $X$ were obtained using Latin Hypercube Sampling (LHS, \cite{mckay2000comparison}) across a six-dimensional space bounded by the respective parameter ranges (see \autoref{t_indep_params}). 
This approach prevents sparse regions in the combustion parameter design space and enables generative learning across the full feature space. The resulting dataset $\mathcal{D}$ contains 1295 points \((x^{(j)}, y^{(j)})\) of parameters and corresponding labels and is visualized in \autoref{fig:Figure_Dataset}. 

\section{Generative Models}\label{sec:Theory}
This Section provides theoretical foundations for the generative models benchmarked in this work. Flow-based generative models learn a target probability distribution by transforming a simple, known source distribution via a series of invertible transformations. The first applications of this concept were introduced by Normalizing Flows (NF, \cite{rezende2015variational}, \cite{dinh2014nice}), which were further developed by Invertible Neural Networks (INN, \cite{ardizzone2018analyzing}). Both are covered in \autoref{sec:NF_INN}. A more recend flow-based approach are Neural Ordinary Differential Equations (NODE, \cite{chen2018neural}), which, instead of relying on a discrete series of mappings, model the transformation of distributions by a time-continuous ODE that is parameterized by a single neural network. Conditional Flow Matching (CFM, \cite{lipman2022flow}, \cite{tong2023conditional}) further improves this novel concept by providing a framework to efficiently train NODE at scale. Both approaches are described in \autoref{sec:NODE_CFM}.\\
An alternative to flow-based models is presented by the well known Generative Adversarial Networks (GAN) which were initially introduced by \cite{goodfellow2020generative}. Emulating a game theoretic two-player setup, a generator network is trained to transform source distribution samples into target distribution samples, while a discriminator network is tasked with identifying generated samples from real samples given some set of original data. GAN, as well as their further development into the Conditional Wasserstein GAN used in this work, are covered in \autoref{sec:CWGAN}  

\subsection{Normalizing Flows and Invertible Neural Networks}\label{sec:NF_INN}
Normalizing flows provide a robust framework for the generative modeling of  complex probability distributions $p_X$ of underlying data $X \in \mathbb{R}^D$. These models are trained to transform $p_X$ into a simple base distribution, typically a standard Gaussian $Z\sim p_Z$ via a sequence of invertible transformations $f=f_1\circ\cdots\circ f_n: \mathbb{R}^D \to \mathbb{R}^D$, such that $Z= f(X)$. With $g$ denoting the inverse of $f$, the complex density can then be modeled by using the change-of-variables formula:
\begin{equation}\label{eq_change_of_vars}
    p_X(x)=p_Z(f(x))\lvert  \det \mathrm{D}f(x)\rvert
    =p_Z(f(x))\lvert  \det \mathrm{D}g(f(x))\rvert^{-1}.
\end{equation}
Here, $D$ denotes the Jacobian matrix. To ensure invertibility and efficient computation of Jacobian determinants, specific architectures such as affine coupling layers \cite{dinh2016density} are implemented as transformations $f_i$. These transformations rely on splitting an input $x\in\R^d$ into halves $u_1=x_{1:k},u_2=x_{k+1:D}$ and constructing the output $v=[v_1,v_2]$ by applying simple transformations that only rely on one half of the input. For affine coupling blocks this transformation is as follows:
\begin{equation}\label{eq_couplingblock_forw}
	\begin{split}
		v_1=u_1\cdot \exp(s_2(u_2))+t_2(u_2),\\ v_2=u_2\cdot \exp(s_1(v_1))+t_1(v_1).
	\end{split}
\end{equation}
In \autoref{eq_couplingblock_forw}, the transformations $s_i,t_i$, $i\in\{1,2\}$ are usually represented by simple feedforward neural networks. By designing the transformations in this way, all $f_i$ have triangular Jacobians, thus $\det Df(x)$ is easy to compute in \autoref{eq_change_of_vars}.

The invertible architecture and tractable Jacobians allow for maximum-likelihood training.
Given a dataset $\mathcal{D}=(x^{(j)})_{j=1}^k$ of samples from $p_X$, the log-likelihood of $\mathcal{D}$ under $f$ given the learned parameters $\theta$ is given by
\begin{equation}\label{eq_maxlikelihood}
    \begin{split}
    \log p(\mathcal{D}\rvert \theta)= \sum_{j=1}^k \log p_X(x^{(j)}\rvert\theta)
    = \sum_{j=1}^k \log p_Z(f(x^{(j)}\rvert \theta) + \log \lvert \det \mathrm{D}f(x^{(j)}\rvert \theta)\rvert.
\end{split}
\end{equation}
Invertible Neural Networks (INN) employ the same class of invertible architectures as normalizing flows, but extend their scope of application and training methodology. Rather than merely approximating an unconditional data distribution $p_X$, INNs aim to learn the full posterior distribution $p_{X \vert Y}(x \vert y)$ of input variables $X$ conditioned on observations $Y \in \mathbb{R}^L$ that are generated by a forward process $s : X \to Y$ and typically follow another complex distribution $p_Y$.

In contrast to normalizing flows, INNs are commonly trained alternately in both the forward (or ``normalizing'') and backward (or ``generative'') directions. During the forward pass, samples $X \in \mathbb{R}^D$ are mapped to a concatenated representation $[Y, Z] = [f_Y(X), f_Z(X)]$, consisting of the corresponding label vector $Y$ and a latent variable $Z \in \mathbb{R}^{D-L}$ drawn from a simple prior distribution $p_Z$ as before. During the backward pass, pairs of labels and latent variables $(Y, Z)$ are mapped back to input parameters $X$, which are expected to reproduce the prescribed output $Y$ under the process $s$.

Instead of relying on the likelihood-based training objective in \autoref{eq_maxlikelihood}, which is commonly used for normalizing flows, INNs are typically trained using a combination of supervised and unsupervised loss terms. Following \cite{ardizzone2018analyzing}, and as adopted in this work, the forward loss $\mathcal{L}_{\mathrm{Forw}}$ and the backward loss $\mathcal{L}_{\mathrm{Rev}}$ are defined as
\begin{eqnarray}
\label{eq_l_forw}
   \mathcal{L}_{\mathrm{Forw}}=\lambda_Y\mathrm{MSE}(f_Y(x^{(j)}),y)+\lambda_Z\mathrm{MMD}([f_Y(x^{(j)}),f_Z(x^{j)})],[y^{(j)},z^{(j)}]).\label{eq:inn_l_forw_2}
\end{eqnarray}
and
\begin{equation}
\label{eq_l_rev}
\mathcal{L}_{\mathrm{Rev}}
= \lambda_X \, \mathrm{MMD}\bigl(g(y^{(j)}, z^{(j)}), x^{(j)}\bigr).
\end{equation}

Here, $(x^{(j)}, y^{(j)})$ denote paired samples from a dataset $\mathcal{D}$, which may be obtained by simulating the forward process $s$, as described for example in \autoref{sec:dataset}, while $z^{(j)}$ represents a sampled noise vector. In \autoref{eq_l_forw}, the Mean Squared Error (MSE) term enforces accurate prediction of the labels $y^{(i)}$, whereas the Maximum Mean Discrepancy (MMD, \cite{gretton2006kernel}) is used in both directions to align the empirical distributions of generated and target samples. MMD is a kernel-based statistical distance that measures the maximum discrepancy between expectations over functions in a reproducing kernel Hilbert space. In both \autoref{eq_l_forw} and \autoref{eq_l_rev}, scalar hyperparameters $\lambda_X,\lambda_Y$ and $\lambda_Z$ are applied such that the respective loss terms remain within a similar order of magnitude during training.  

Once sufficiently trained, an INN can be employed in the generative direction to produce diverse input candidates $x$ that result in a prescribed output $y$ under the considered process $s$. For the combustor dataset introduced in \autoref{sec:dataset}, an extensive investigation of the applicability of INNs is presented in \cite{krueger2025generative}.

Despite their ability to model both conditional and unconditional distributions with high accuracy, both normalizing flows and INNs may suffer from practical limitations. In particular, the computation of inverses and Jacobian determinants can be computationally expensive and numerically unstable. Even when training objectives are formulated to avoid explicit Jacobian evaluations, model expressiveness may remain constrained due to the requirement of specialized invertible architectures, such as those defined in \autoref{eq_couplingblock_forw}.

\subsection{Neural ODE and Conditional Flow Matching}\label{sec:NODE_CFM}
While INN and NF apply "discrete" invertible transformations $f_1\circ\cdots\circ f_n$, Neural Ordinary Differential Equations (NODE, \cite{chen2018neural}) introduce a continuous-time approach by treating the transformation as a dynamical system that is expressed as an ODE defined by a vector field $v$ that is represented by a single neural network with parameters $\theta$:
\begin{equation}\label{eq_node}
    \frac{dx(t)}{dt}=v(x(t),t,y,\theta).
\end{equation}
Defining input data $x(0)=x_0=\phi_0(x)$ as initial condition to \autoref{eq_node}, the output of the NODE is defined as the solution $x(1)=x_1=\phi_1(x)$ at time $t=1$. Here, $\phi_t(x)$ is the solution map or \textit{flow} of the NODE that transports $x$ from time $t=0$ to time $t$ along the vector field $v$:
\begin{equation}\label{eq_node_flow}
    x_1=\phi_1(x)=x_0+\int_0^1 v(x_t,t,y,\theta)\approx S(x_0,v,t_0,t_1,\theta)
\end{equation}
In \autoref{eq_node_flow}, $S$ denotes a numerical ODE solver used to approximate the solution in practice.  
Following the notation from \autoref{sec:NF_INN}, NODE can be applied to model complex distributions $p_X$ by choosing $p_1=p_X, p_0=p_Z$ and full posteriors $p_{X\vert Y}(x\vert y)$ by setting $Y\sim p_Y$ as an additional conditioning input for $v$ in \autoref{eq_node}. $\phi_t$ then induces a \emph{probability path} $p_t:[0,1]\times\R^d\to\R^d$ with $p_0=p_Z$ and $p_1=p_X$ via the pushforward operation:
\begin{equation}\label{eq:prob_path}
    p_t(x)
    =
    ([\phi_t]_*p_0)(x)
    = 
    p_0(\phi_t^{-1}(x)) \left| \det \frac{\partial\phi_t^{-1}}{\partial x}(x) \right|.
\end{equation}
In \autoref{eq:prob_path}, $p_t$ is said to be \textit{generated} by the vector field $v$ from \autoref{eq_node}.
The resulting \textit{Continuous Normalizing Flow (CNF)} provide several advantages over INN: First, the network representing the vector field $v$ does not need to be invertible. This is due to the inverse being computed simply by integrating backwards in time, which, opposed to INN, takes the same computational effort as the forward pass. Furthermore, as proven in \cite{chen2018neural}, the time-continuous approach of CNF simplifies the change-of-variables Formula (\autoref{eq_change_of_vars}) to 
\begin{equation}\label{eq_cont_cow}
    \frac{\partial\log p(x(t))}{\partial t}=-\mathrm{tr}\frac{dv}{dx(t)},
\end{equation}
reducing the log determinant operation to a trace operation. This reduces computational cost to scale linearly with the depth of $v$ and thus makes likelihood-based training approaches more viable in practice. However, training CNF now requires backpropagating through a numerical ODE solver, which is usually done via the \textit{adjoint sensitivity method} (see \cite{chen2018neural}, \cite{pontryagin2018mathematical}) that solves additional augmented ODE based on \autoref{eq_node} to calculate the required gradient $dx/d\theta$. As this requires frequent calls to a sufficiently precise ODE solver during each training iteration, scalability to high dimensional complex data is limited.\\
\ \\
Conditional Flow Matching (CFM, \cite{lipman2022flow},\cite{tong2023conditional}) aims to provide a method of training CNF without requiring calls to an ODE solver by directly regressing the learned vector field $v$ (see \autoref{eq_node}) against a target vector field $u_t$ that generates a desired probability path as per \autoref{eq:prob_path}. This leads to the \textit{Flow Matching Loss}
\begin{equation}\label{eq_fm_loss}
    \mathcal{L}_{\mathrm{FM}}(\theta)=\mathbb{E}_{t, p_t(x)}\Vert v_t(x,\theta)-u_t(x)\Vert^2.
\end{equation}
While \autoref{eq_fm_loss} provides a simple and thus attractive training objective, target vector fields $u_t$ and corresponding probability paths $p_t$ are generally unknown. To solve this, we assume $p_t(x)$ to be a mixture of probability paths $p_t(x\vert c)$ depending on conditions $c\sim p_c$, e.g. 
\begin{equation}\label{eq_pt}
    p_t(x)=\int\ p_t(x\mid c)p_c(c) dc.
\end{equation}
It is shown in \cite{chen2018neural}, Theorem 3.1 that an admissible vector field is then obtained via
\begin{equation}\label{eq_ut}
    u_t(x)=\frac{u_t(x\mid c)p_t(x\mid c)}{p_t(x)},
\end{equation}
where $u_t(x\mid c)$ are conditional vector fields that generate the respective $p_t(x\mid c)$ depending on $c$.
As \autoref{eq_ut} is still difficult to compute due to the integral in \autoref{eq_pt}, the training objective from \autoref{eq_fm_loss} is reformulated to directly match $v_t$ against the conditional vector field $u_t(x\mid c)$ on a per-sample basis:
\begin{equation}\label{eq_loss_cfm}
    \mathcal{L}_{\mathrm{CFM}}(\theta)=\mathbb{E}_{t,p_c(c),p_t(x\mid c)}\Vert v_t(x)-u_t(x\mid c)\Vert^2.
\end{equation}
Theorem 3.2 of \cite{chen2018neural} furthermore shows that \autoref{eq_loss_cfm} still regresses $v_t$ against the marginal vector field $u_t(x)$ from \autoref{eq_ut} by proving $\nabla_{\theta}\mathcal{L}_{\mathrm{FM}}=\nabla_{\theta}\mathcal{L}_{\mathrm{CFM}}$ up to an independent constant. Thus, initially intractable marginal vector fields $u_t(x)$ can be approximated as long as admissible $p_c$, $p_t(x\mid c)$ and $u_t(x\mid c)$ can be efficiently sampled and computed. Although there are several ways of choosing paths $p_t(x\mid c)$ and the corresponding $p_c$, this work applies the simplest form, which defines $p_c=p_Xp_Z=p_1p_0$ as the independent coupling between source and target distributions and sets $p_t$ as the linear interpolation between $x_1$ and $x_0$:
\begin{equation}\label{eq:pt}
    p_t(x\mid c)=tx_1+(1-t)x_0.
\end{equation}
$p_t$ as in \autoref{eq:pt} can then be shown (\cite{chen2018neural}, Proposition 3.3) to be generated by
\begin{equation}
    u_t(x\mid c)=x_1-x_0.
\end{equation}
Thus, the conditional flow matching loss is easily obtained by sampling $x_0=Z\sim p_Z$ and $x_1=X\sim p_X$ independently and computing $p_t$ and $u_t$ as above.

\subsection{Conditional Wasserstein Generative Adversarial Networks}\label{sec:CWGAN}
The structure of a Generative Adversarial Network (GAN) \cite{goodfellow2014gans} is inspired by human learning behavior and is formulated as a game-theoretic two-player optimization problem. In this setting, a generator network is tasked with producing realistic synthetic samples that resemble a target data distribution $X \sim p_X$, while an adversarial discriminator network aims to distinguish between generated samples and real samples drawn from the dataset.

Formally, let $p_X$, $p_Y$, and $p_Z$ be defined as above. We introduce a generator network $G(z; \theta_g)$, parameterized by $\theta_g$, and a discriminator network $D(x; \theta_d)$, parameterized by $\theta_d$. These networks are defined as mappings
\[
G(\cdot; \theta_g) : \mathbb{R}^{d_Z} \rightarrow \mathbb{R}^{d_X},
\qquad
D(\cdot; \theta_d) : \mathbb{R}^{d_X} \rightarrow [0,1],
\]
where the generator transforms latent variables $Z \sim p_Z$ into samples in data space, and the discriminator assigns to each input a scalar value representing the probability that the sample originates from the true data distribution $p_X$.

The resulting adversarial learning process is captured by the following two-player minimax game:
\begin{equation}
\label{eq:Equation_GAN_Minimax}
\min_G \max_D V(D, G)
= \mathbb{E}_{x \sim p_X} \bigl[\log D(x)\bigr]
+ \mathbb{E}_{z \sim p_Z} \bigl[\log \bigl(1 - D(G(z))\bigr)\bigr].
\end{equation}

According to \cite{goodfellow2014gans}, the minimax optimization problem defined in \autoref{eq:Equation_GAN_Minimax} admits a unique solution in which the generator implicitly learns the data distribution, i.e., $G(z; \theta_g) \sim p_X$. Training is performed using an alternating adversarial optimization scheme:
In each iteration, latent samples $z^{(j)} \sim p_Z$ and data samples $x^{(j)} \sim p_X$ are drawn. The discriminator parameters are updated via gradient-based optimization by minimizing an empirical risk that encourages correct discrimination between real and generated samples. Analogously, the generator parameters are updated to minimize an empirical loss that promotes the generation of samples that the discriminator classifies as real. The discriminator and generator losses are given by
\begin{equation}
\begin{aligned}
L_D &= \sum_{j=1}^{m} \left[ \log D(x^{(j)}) + \log \bigl(1 - D(G(z^{(j)}))\bigr) \right], \\
L_G &= \sum_{j=1}^{m} \log \bigl(1 - D(G(z^{(j)}))\bigr).
\end{aligned}
\end{equation}

In \cite{goodfellow2014gans}, it is shown that, for a fixed generator $G$, the maximization of the value function with respect to the discriminator admits a closed-form expression. Denoting by $p_G$ the model distribution implicitly induced by the generator $G$, i.e., the distribution of samples $G(Z)$ with $Z \sim p_Z$, one obtains
\begin{equation}
\label{eq:JS}
\begin{aligned}
\max_D V(D, G)
&= \mathbb{E}_{X\sim p_X,\,Z\sim p_Z}\frac{1}{m}L_D \\
&= -\log(4)
+ \mathrm{KL}\!\left( p_X \,\middle\|\, \frac{p_X + p_G}{2} \right)
+ \mathrm{KL}\!\left( p_G \,\middle\|\, \frac{p_X + p_G}{2} \right) \\
&= -\log(4) + \mathrm{JS}(p_X, p_G).
\end{aligned}
\end{equation}

This result implies that the minimax game reaches its global optimum if and only if $\mathrm{JS}(p_X, p_G) = 0$, which is equivalent to $p_G = p_X$.

Under a set of idealized assumptions, it can be shown that the generator distribution $p_G$ converges to the data distribution $p_X$. These assumptions include, among others, monotonic improvements of the generator with respect to the training objective in \autoref{eq:Equation_GAN_Minimax} at each iteration, as well as the ability to train the discriminator to optimality for any fixed generator configuration.

In practice, however, these conditions are difficult to satisfy or verify. As a consequence, GAN training is often unstable and prone to issues such as mode collapse, as discussed in \cite{salimans2016modecollapse}. This typically necessitates extensive hyperparameter tuning, careful initialization, or the use of specialized GAN variants. A prominent remedy to these shortcomings is provided by Wasserstein GANs, which are discussed in \autoref{sec:cwgan}.

\subsubsection{Conditional Gan}
\label{sec:cGAN}
Conventional GANs, as described above, do not provide explicit control over the generated outputs. However, as already noted in \cite{goodfellow2014gans}, the framework can be extended to model conditional distributions. This idea was formalized in subsequent work on Conditional GANs (cGANs) \cite{mirza2014conditionalgans}, which we briefly review here.

Let $p_X$ be defined as before, and let $Y \sim p_Y$ denote labels associated with data points, as introduced in \autoref{sec:NF_INN}. In this conditional setting, the learning objective is no longer the marginal data distribution but rather the posterior distribution $p_{X \mid Y}(x \mid y)$. By providing the conditioning variable $Y$ as an additional input to both the generator and the discriminator, the adversarial optimization problem becomes
\begin{equation}
\label{eq:Loss_CGAN}
\min_G \max_D V(D, G)
= \mathbb{E}_{x \sim p_X} \bigl[\log D(x \mid y)\bigr]
+ \mathbb{E}_{z \sim p_Z} \bigl[\log\bigl(1 - D(G(z \mid y))\bigr)\bigr].
\end{equation}

According to \autoref{eq:Loss_CGAN}, the discriminator is trained to distinguish between samples drawn from the true joint distribution $(X, Y) \sim p_{(X,Y)}$ and samples from the joint distribution induced by the generator, given by $(G(Z, Y), Y) \sim p_{(G_\theta(Z,Y), Y)}$.

In contrast to the unconditional case, the discriminator therefore evaluates not only whether a sample is realistic, but also whether it is consistent with the provided condition $Y$. Simultaneously, the generator learns to produce samples that are both indistinguishable from real data and agree with the specified conditioning variables. In this way, cGANs enable controlled and conditional sample generation.

\subsubsection{Conditional Wasserstein Gan}\label{sec:cwgan}

Wasserstein GANs (WGANs), introduced by \cite{arjovsky2017wassersteingan}, address key shortcomings of conventional GANs—most notably mode collapse \cite{kossale2022mode}—by replacing the Jensen--Shannon divergence in \autoref{eq:JS} with the Earth Mover’s (Wasserstein-1) distance defined as
\begin{equation}
    W(p_X, p_G)
= \inf_{\gamma \in \Pi(p_X, p_G)}
\mathbb{E}_{(x,y)\sim\gamma}\bigl[\|x-y\|\bigr].
\end{equation}
Here, $\Pi(P_X, P_G)$ denotes the set of all joint distributions $\gamma(x,y)$ whose marginals are $p_X$ and $p_G$, respectively.

By the Kantorovich--Rubinstein duality \cite{villani2008optimal}, the Wasserstein-1 distance admits the equivalent dual formulation
\begin{equation}
\label{eq:Kanto-Rubinstein_eq}
W(p_X, p_G)
= \sup_{\|f\|_L \le 1}
\left(
\mathbb{E}_{x \sim p_X}[f(x)]
- \mathbb{E}_{x \sim p_G}[f(x)]
\right),
\end{equation}
where the supremum is taken over all 1-Lipschitz continuous functions $f$.

A key advantage of the Wasserstein distance over the Jensen--Shannon and Kullback--Leibler divergences is its continuity and almost-everywhere differentiability with respect to the generator parameters, provided mild regularity assumptions on the push-forward map $g_\theta$ inducing $p_G$ are satisfied. In contrast, analogous continuity properties do not generally hold for the Jensen--Shannon divergence.

Direct computation of the Wasserstein distance is typically intractable. In practice, it is approximated by introducing a critic network $f_w$, which replaces the discriminator used in standard GANs and aims to approximate the optimal 1-Lipschitz function. The critic is trained to maximize
\begin{equation}
\label{eq:Critic_criterium}
\mathcal{L}_{f_w}
= \mathbb{E}_{x \sim p_X}[f_w(x)]
- \mathbb{E}_{x \sim p_G}[f_w(x)],
\end{equation}
while the generator is trained by minimizing
\[
\mathcal{L}_G
= - \mathbb{E}_{x \sim p_G}[f_w(x)].
\]

The WGAN framework has been shown to substantially improve training stability and to significantly mitigate mode collapse, thereby providing a more robust alternative for adversarially trained generative models. Moreover, the formulation extends naturally to the conditional setting: In this case, the Wasserstein-1 distance between conditional distributions is defined as
\begin{equation}
   W(p_{X\mid Y}(x \mid y), p_G(x \mid y))
= \inf_{\gamma \in \Pi(p_{X\mid Y}(x \mid y), p_G(x \mid y))}
\mathbb{E}_{(x,y)\sim\gamma}\bigl[\|x-y\|\bigr], 
\end{equation}
where $P_{X\mid Y}(x \mid y)$ and $P_G(x \mid y)$ denote the conditional distributions of real and generated data given the condition $y$, respectively.

Applying the Kantorovich--Rubinstein duality and approximating the resulting supremum by a conditional critic network $f_w$, the conditional WGAN objective can be written as
\begin{equation}
    \min_G \max_{f_w}
\mathbb{E}_{x \sim P_{X\mid Y}(x \mid y)} \bigl[f_w(x,y)\bigr]
- \mathbb{E}_{z \sim p_Z(z)} \bigl[f_w(G(z,y),y)\bigr],
\end{equation}
where $f_w(x,y)$ denotes a parameterized conditional critic and $G(z,y)$ is the generator output corresponding to latent variable $z$ and condition $y$.

\subsection{Bayesian Inference for Posterior Sampling in Inverse Problems}

As before, we consider an inverse problem in which the objective is to infer an unknown parameter
$x \in \mathbb{R}^D$ from observations $Y \in \mathbb{R}^L$, which are assumed to arise through a
forward model $s : \mathbb{R}^D \to \mathbb{R}^L$ that is subject to additive noise, se \eqref{eqa:regression}.
Under this assumption, the likelihood function
$p_{Y \mid X} : \mathbb{R}^L \times \mathbb{R}^D \to \mathbb{R}_+$ is given by
\begin{equation}
    p_{Y \mid X}(y \mid x)
    \propto
    \exp\!\left(
        -\frac{1}{2\sigma^2}
        (y - s(x))^\top (y - s(x))
    \right).
\end{equation}
Given a prior density $p_X(x)$ encoding a priori information about the parameter,
Bayes’ theorem yields the posterior distribution
\begin{equation}
    p_{X \mid Y}(x \mid y)
    \propto
    p_{Y \mid X}(y \mid x)\, p_X(x).
\end{equation}

In many applications of interest, particularly in engineering and the natural sciences,
the evaluation of the forward model $s(x)$ is computationally expensive, as it may
require the numerical solution of complex physical models. Consequently, Markov Chain Monte Carlo (MCMC) methods become computationally prohibitive.

To alleviate this computational bottleneck, a surrogate-based approach, realized by e.g. a simple feedforward neural network, replaces the
forward model $s$ by a computationally inexpensive approximation
$f : \mathbb{R}^D \to \mathbb{R}^L$.
This induces a surrogate likelihood
\begin{equation}
    p_{Y \mid X,S}(y \mid x)
    \propto
    \exp\!\left(
        -\frac{1}{2\sigma^2}
        (y - f(x))^\top  (y - f(x))
    \right),
    \label{eq:conditional_density}
\end{equation} where \(\Sigma^{-1}:={I_L (\tau^2+\omega^2)}^{-1}\). The resulting surrogate-based posterior density is therefore given by
\begin{equation}
    p_{X \mid Y,S}(x \mid y)
    \propto
    p_{Y \mid X,S}(y \mid x)\, p_X(x).
\end{equation}
This distribution serves as a computationally tractable approximation to the true
posterior $p_{X \mid Y}(x \mid y)$.
Sampling from $p_{X \mid Y,S}(x \mid y)$ may be performed using a Metropolis-algorithm. Given a proposal $\tilde{x}_t$ generated from a proposal distribution
$q(\tilde{x}_t \mid x_t)$, the acceptance probability is given by
\begin{equation}
    \rho(x_t, \tilde{x}_t)
    =
    \min\left\{
    1,\,
    \frac{
        \exp\!\left(
        -\frac{1}{2\sigma^2}
        (y - S(\tilde{x}))^\top  (y - f(\tilde{x}))
    \right)
    }{
        \exp\!\left(
        -\frac{1}{2\sigma^2}
        (y - S(x))^\top  (y - f(x))
    \right)
    }
    \right\}.
\end{equation}

The diagonal entries of the covariance matrix \(\Sigma\) can alternatively be treated as
hyperparameters rather than being estimated empirically from surrogate residuals,
reflecting the epistemic nature of the surrogate approximation error.

Provided that the proposal distribution satisfies standard irreducibility and
aperiodicity conditions (e.g.\ a Gaussian random-walk proposal), the resulting Markov
chain is ergodic and converges in distribution to the surrogate-based posterior
$p_{X \mid Y,S}(x \mid y)$.
As an alternative to the fully surrogate-based Metropolis algorithm, one may
adopt a Delayed Acceptance Metropolis scheme
\cite{banterle2015delayedmetropolis}, in which surrogate-based likelihood
evaluations provide a computationally inexpensive first-stage filter, and only
accepted proposals are subsequently evaluated using the full numerical forward
model.

\section{Implementation}\label{sec:implementation}
This section provides implementation details of all models used in this work. Surrogate models discussed in \autoref{sec:su_data_aug} were employed to quickly validate generative models, as well as to create larger training datasets. Architecture and training hyperparameters of all generative models compared in this work are covered in \autoref{sec:implementation_gen_models}
\subsection{Surrogate Models and Data Augmentation}\label{sec:su_data_aug}
The benchmarking experiments covered in \autoref{sec:Validation} require that performance labels are obtained for a large number of generated designs. As the simulation of a single design $x$ (see \autoref{sec:dataset}) takes about 96 core hours, this approach becomes infeasible. Instead, three distinct surrogate models denoted as \( S_{U_M} \), \( S_{\p} \) and \( S_G \) were trained to predict performance labels for design parameters $X$:

\begin{equation}\label{eq:su_pred}
 Y^S= (Y_i^S)_{i\in\{1,2,3\}}= (U^S_{M}, \p^S, G^S) = (S_{U_M}(X), S_{\p}(X), S_G(X)).   
\end{equation}

Each surrogate model was implemented as a multilayer perceptron comprising five fully connected layers, with layer widths reaching 200 neurons. ReLU functions served as non-linear activations between layers. Model training was conducted using the Adam optimization algorithm on 1{,}000 samples drawn from the original dataset $\mathcal{D}$ defined in \autoref{sec:dataset}. The remaining $n=295$ points of $\mathcal{D}$ were used to evaluate the performance of the surrogate models using the mean absolute error between predicted and actual label values: 

\begin{equation}\label{eq:su_err_rel}
\epsilon^S_{Y_i} = \frac{1}{n} \sum_{j=1}^{n} \lvert y^{(j),S}_i - y_i^{(j)}\rvert\,.
\end{equation}
 
 The resulting errors (\autoref{tab:mae_errors}) demonstrate that the surrogate models provide a reliable approximation of the full simulation workflow.
\begin{table}[h!]
\centering
\caption{Mean absolute surrogate model errors for each label}
\label{tab:mae_errors}
\begin{tabular}{lccc}
\toprule
Label  & \( U_M \) & \( \Delta p_{t,\text{rel}} \) & \( G \) \\
\midrule
\( \epsilon^S_{Y_i} \) & 0.0131 & 0.00014 & 0.0251 \\
\bottomrule
\end{tabular}
\end{table}

To eliminate the potential error introduced by the surrogate models, as well as to investigate the effects of larger training datasets, a synthetic dataset $\mathcal{D}_{\mathrm{Aug}}=(x^{(j)},y^{(j),S})_{j=1}^n$ containing $n=250\,000$ points was created. Parameter vectors $X$ were sampled by LHS as in \autoref{sec:dataset}. Instead of obtaining the corresponding labels via the costly simulation workflow, they were predicted as in \autoref{eq:su_pred}. This augmented dataset and the surrogate models were used to perform all benchmarking experiments of generative models described in the following.

\subsection{Generative Models}\label{sec:implementation_gen_models}
Like the surrogate models in \autoref{sec:su_data_aug}, the underlying neural networks of all generative models were implemented as fully connected multilayer perceptrons with varying widths and depths and were trained within the \texttt{pytorch} \cite{paszke2017automatic} Framework in \texttt{Python3} using and Adams \cite{adam2014method} optimizer. Furthermore, one model of each generative architecture was trained on the dataset sizes  
\begin{equation}\label{eq_datasizes}
d=\lvert \mathcal{D}_{\mathrm{Aug}}\rvert\in\{100, 500, 1000, 5000, 10000, 50000, 100000\}.    
\end{equation}
\subsubsection{Conditional Wasserstein GAN Implementation}
\label{sec:Implementation_cWGAN}

The generator architecture comprises five fully connected layers of $1500$ neurons. Again, ReLU activations are used between hidden layers. A sigmoid activation is applied to the five continuous output variables to constrain them to the interval \([0,1]\). The categorical parameter $N_H$ is produced via a softmax activation to ensure a valid probability distribution across its possible categories.\\
The critic network consists of three fully connected layers of $64$ neurons. Leaky ReLU activation functions \cite{xu2020reluplex} with a negative slope of $0.2$ are applied to the hidden layers, while the output layer uses the identity function to comply with the Wasserstein loss formulation.

To enforce the Lipschitz continuity constraint required by the Wasserstein GAN, a gradient penalty term is incorporated into the loss function of the critic:
\begin{equation}\label{eq:penalty}
L_{\text{grad-penalty}}(\tilde{x},y) = \lambda \frac{1}{n} \sum_{j=1}^{n} \left( \| \nabla_{\tilde{x}^{(j)}} f_w(\tilde{x}^{(j)},y^{(j)}) \|_2 - 1 \right)^2.
\end{equation}
in \autoref{eq:penalty}, \(\tilde{x}\) denotes samples linearly interpolated between real and generated data, \(y\) denotes the corresponding condition and \(\lambda = 10\) is the regularization coefficient.

All generators receive 150{,}000 gradient training updates. For each generator update, the critic is updated five times to ensure stability of the gradients used for generator training. The optimizer used a learning rate of $0.0002$ and a modified $\beta$ parameter of $(0.0, 0.9)$. 

After each epoch, the generator is tested on a fixed test set of $n=1\,000$ samples $(x^{(j)},y^{(j)})$: A design $x^{(j),\mathrm{Gen}}$ is generated for each target vector $y^{(j)}$ and the mean squared error (defined analogously to \autoref{eq:mae}) between true label values $y_i^{(j),\mathrm{Gen}}=S_{y_i}(x^{(j),\mathrm{Gen}})$ obtained by the surrogate models (see \autoref{sec:su_data_aug}) and initial target values for each label component $y_i^{(j)}$ of $y^{(j)}$ is computed for $y_i\in\{U_M,\p,G\}$.

Over the course of training, the model achieving the lowest overall validation loss is retained.
For all dataset sizes, the best possible models were selected using grid search over different batch sizes. The results are shown in \autoref{t:batchsizes_INN_CFM_WGAN}.

\subsubsection{Flow-Based Models}
INN were implemented using the Framework for Easily Invertible Architectures (\texttt{FrEIA}) in Pytorch. All models consisted of $10$ affine coupling blocks as described by 
\autoref{eq_couplingblock_forw}, where the networks representing the subfunctions $s_i, t_i$ consisted of two layers of $115$ neurons. Weights $\lambda_X$, $\lambda_Y$ and $\lambda_Z$ in Equations \ref{eq_l_forw} and \ref{eq_l_rev} were set to $20,40$ and $4$, respectively. \\CFM Models representing the vector field $v$ in \autoref{eq_loss_cfm} consisted of five layers of $500$ neurons. Between hidden layers, SeLU activation functions \cite{klambauer2017self} were applied. The \texttt{torchcfm} library provided by \cite{tong2023conditional} for Pytorch was used to the conditional probability paths, vector fields and the flow matching loss from Equations \ref{eq_pt}, \ref{eq_ut} and \ref{eq_loss_cfm}, respectively. \\
Both CFM and INN models were trained for $4800$ total epochs with an initial learning rate of $0.001$ that was reduced by a factor of $10$ after $1600$ and $3200$ epochs. Depending on the training dataset size, optimal results were obtained for CFM and INN models by varying batch sizes, which are listed in  \autoref{t:batchsizes_INN_CFM_WGAN}.

\begin{table}[]
    \centering
    \begin{tabular}{c|ccccccc}
    \toprule
        $\lvert \mathcal{D}_{\mathrm{Aug}}\rvert$ &$100$& $500$& $1000$& $5000$& $10000$& $50000$& $100000$  \\
        \hline
         INN&5&5&20&50&500&1000&1000\\
         CFM&50&100&100&500&500&2500&5000\\
         WGAN&50&50&50&500&500&1000&1000\\
    \bottomrule
    \end{tabular}
    
    \caption{Training batch sizes for INN and CFM models depending on training dataset sizes $\lvert\mathcal{D}_{\mathrm{Aug}}\rvert$.}
    \label{t:batchsizes_INN_CFM_WGAN}
\end{table}

\subsubsection{Bayesian Inference}
The proposal distribution was chosen as a Gaussian centered at the current state of the Markov chain, with a standard deviation of $0.025$ for each continuous variable. To ensure feasibility, proposals were truncated to the hypercube $[0,1]^5$. The covariance matrix $\Sigma$ appearing in \autoref{eq:conditional_density} was assumed to be diagonal, and its diagonal entries were treated as hyperparameters and set to $1.75 \times 10^{-6}$.

For the single categorical variable $N_H$, proposals were generated only for the continuous subvector $X_{\mathrm{cont}}$. The discrete component was then determined by evaluating all admissible values of $N_H$ in combination with the proposed $X_{\mathrm{cont}}$. Among these candidates, the value of $N_H$ that minimized the mean squared error (cf. \autoref{eq:mae}) between the surrogate-predicted labels $Y^S$ and the target labels $Y$ was selected.

Each Markov chain was executed for a fixed number of iterations, with the first $10{,}000$ samples discarded as burn-in to mitigate initialization bias. The final sample set was obtained by retaining the accepted samples from each chain after the burn-in period.

\subsubsection{Surrogate Models for Bayesian Inference}
Each surrogate model was implemented as a fully connected feedforward neural network in PyTorch. Three independent networks were trained in parallel, one for each component of the target vector~$Y$, see \autoref{eq:su_pred}. The architecture consists of an input layer of dimension~10, a hidden layer of width~64 followed by six additional hidden layers of the same width, and a single-output linear layer. All hidden layers use a LeakyReLU activation function with slope of $\alpha = 0.2$.

For each configuration, a fixed test set of 1{,}000 samples distinct the training data was employed for model selection.

Training was performed using the Adam optimizer with a learning rate of $10^{-3}$ and momentum parameters $(\beta_1,\beta_2) = (0.5, 0.999)$. The mean squared error between surrogate predictions and reference labels was used as the training objective.

A hyperparameter search was conducted independently for each training configuration to identify optimal batch sizes in the range of 5 to 1000. For each case, the surrogate models achieving the lowest MSE on the test set were retained. All surrogate models were trained using approximately $4 \times 10^4$ gradient update steps.

\section{Validation}\label{sec:Validation}
This section evaluates the performance of the proposed generative approaches with respect to both accuracy and diversity in solving the inverse design task. Two complementary studies are performed: The first (\autoref{sec_val_accuracy}) assesses the accuracy of generated designs relative to prescribed target labels, while the second (\autoref{sec_val_diversity}) analyzes the diversity and robustness of generated solutions under fixed target conditions. 
\begin{figure}[!htb]
    \centering
    \includegraphics[width=\linewidth]{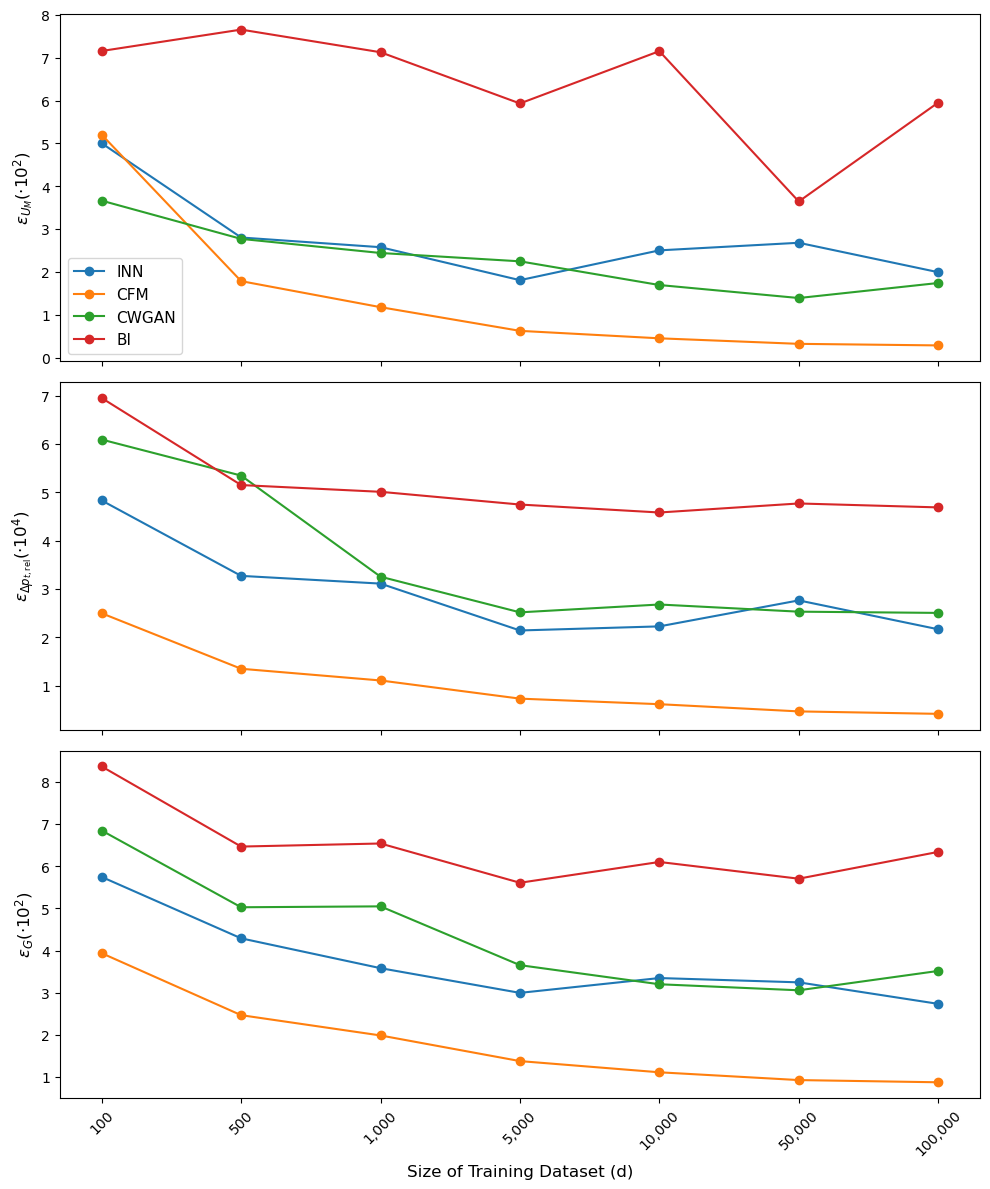}
    \caption{Mean absolute errors between target label values and true label values of generated designs for all dataset sizes $d$ and labels $Y_i\in\{U_M,\p,G\}$.  }
    \label{fig:plot_mae}
\end{figure}

\subsection{Accuracy}\label{sec_val_accuracy}
The accuracy of the generative approaches was compared depending on the training dataset size $d$ (see \autoref{eq_datasizes}) for a wide variety of targets. Target vectors $y^{(j)}$ with target label values $y_i^{(j)}\in\{U_M,\p,G\}$ were taken from a fixed test subset $\mathcal{D}_{\mathrm{Test}}=(x^{(j)},y^{(j)})_{j=1}^N$ of $\mathcal{D}_{\mathrm{Aug}}$ comprising $n=1000$ parameter-label pairs. A design Vector $x^{(j),\mathrm{Gen}}$ was generated for each target vector by all trained models. Corresponding true label values $y_i^{j,\mathrm{Gen}}=S_{y_i}(x^{(j),\mathrm{Gen}})$ were obtained using the surrogate models. The mean absolute error (\autoref{eq:mae}) between target labels $y_i^{(j)}$ and $y_i^{(j),\mathrm{Gen}}$ is given for all models and dataset sizes in \autoref{fig:plot_mae} and \autoref{tab_accuracy}.

\begin{equation}
    \label{eq:mae}
    \epsilon_{Y_i} = \frac{1}{n}\sum_{j=1}^n \left | y^{(j)}_i - y_i^{(j),\mathrm{Gen}}\right |=\frac{1}{n}\sum_{j=1}^n \left | y^{(j)}_i - S_{Y_i}(x^{(j),\mathrm{Gen}})\right |\,.
\end{equation}

\begin{table}[h]
\caption{Mean absolute errors between target label values and groundtruth surrogate predictions for generated designs as per \autoref{eq:mae}}.
\vspace{0.5mm}
\centering
\label{tab_accuracy}
\begin{tabular}{c c |c c c c c c c}
\toprule
 & \multicolumn{8}{c}{$d$} \\ 
\cmidrule(lr){3-9}
\textbf{Label}& \textbf{Model} & 100 & 500 & 1000 & 5000 & 10000 & 50000 & 100000\\
\midrule

\multirow{4}{*}{$U_M$} 
  & INN       & 0.0500 & 0.0281 & 0.0258 & 0.0181 & 0.0251 & 0.0268 & 0.0200 \\
  & CFM       & 0.0520 & 0.0179 & 0.0118 & 0.0063 & 0.0045 & 0.0033 & 0.0029 \\
  & CWGAN     & 0.0366 & 0.0277 & 0.0244 & 0.0225 & 0.0170 & 0.0139 & 0.0174 \\
  & BI        & 0.0716 & 0.0765 & 0.0713 & 0.0593 & 0.0715 & 0.0365 & 0.0595 \\
  
\midrule

\multirow{4}{*}{$\p$} 
  & INN       & 0.00048 & 0.00033 & 0.00031 & 0.00021 & 0.00022 & 0.00028 & 0.00022 \\
  & CFM       & 0.00025 & 0.00014 & 0.00011 & 0.00007 & 0.00006 & 0.00005 & 0.00004 \\
  & CWGAN     & 0.00061 & 0.00053 & 0.00033 & 0.00025 & 0.00027 & 0.00025 & 0.00025 \\
  & BI        & 0.00070 & 0.00052 & 0.00050 & 0.00048 & 0.00046 & 0.00048 & 0.00047 \\

\midrule

\multirow{4}{*}{$G$} 
  & INN       & 0.0575 & 0.0429 & 0.0358 & 0.0300 & 0.0335 & 0.0325 & 0.0274 \\
  & CFM       & 0.0394 & 0.0247 & 0.0199 & 0.0138 & 0.0111 & 0.0093 & 0.0087 \\
  & CWGAN     & 0.0685 & 0.0503 & 0.0505 & 0.0365 & 0.0320 & 0.0306 & 0.0352 \\
  & BI        & 0.0837 & 0.0647 & 0.0654 & 0.0561 & 0.0610 & 0.0571 & 0.0634 \\

\bottomrule
\end{tabular}
\end{table}
\begin{figure}[!htb]
    \centering
    \includegraphics[width=\textwidth]{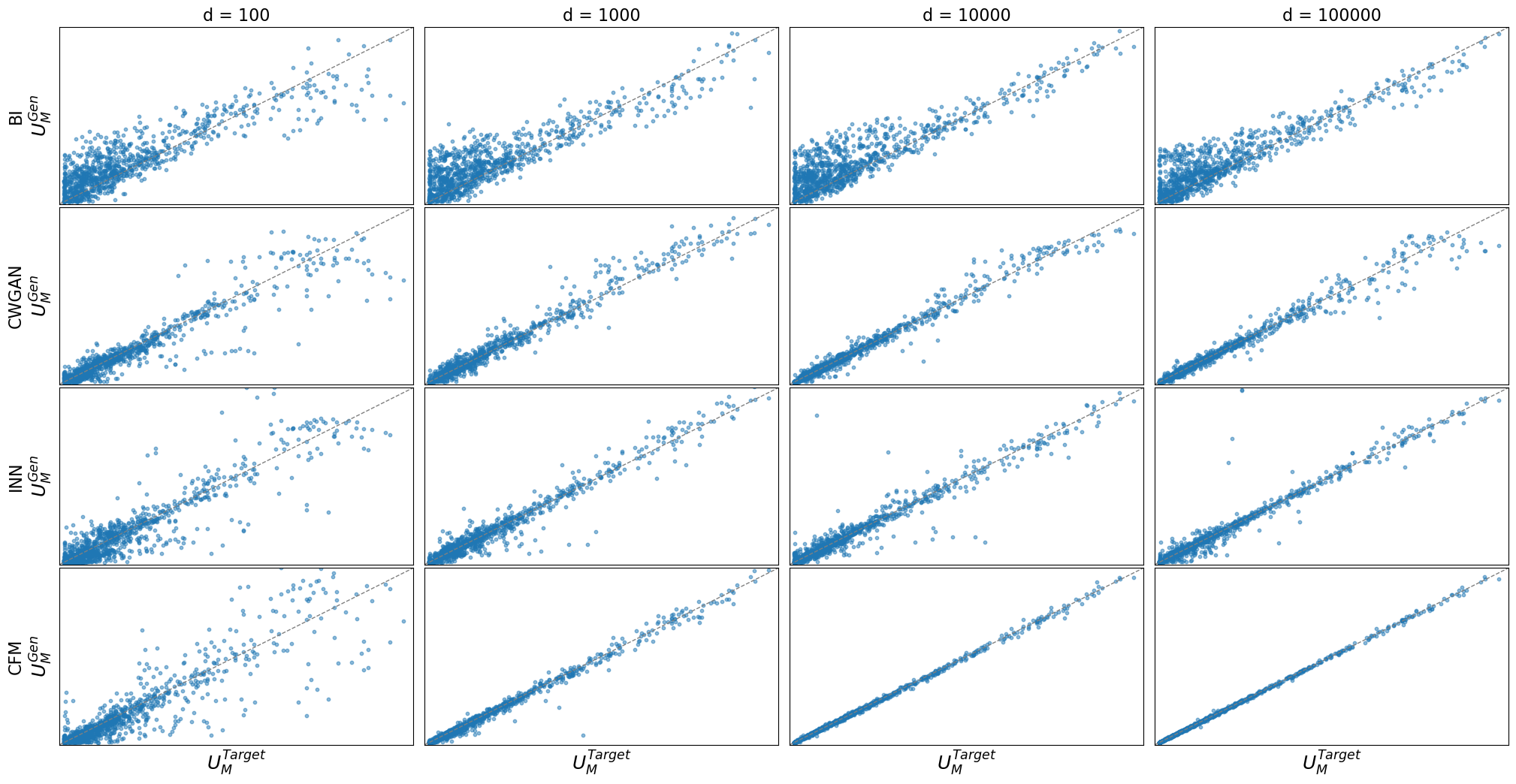}
    \caption{Parity plots between target and true label values for $Y_i=U_M$ and $d\in\{100,1\,000, 10\,000, 100\,000\}$.}
    \label{fig:parity_plot_comparison_u_m_limited_window}
\end{figure}

The best performance is achieved by the CFM models on all training dataset sizes and labels except for $Y_i=U_M$ at $d=100$. Relative improvements over the next best respective model range from $31.5$ to $83.3$ percent depending again on dataset size and label. It can furthermore be observed that CFM models still improve significantly on training datasets  sizes over $d=5\,000$, while the other approaches tend to stagnate or even deteriorate. 

\autoref{fig:parity_plot_comparison_u_m_limited_window} illustrates differences in the quality of generated design vectors $x^{(j),\mathrm{Gen}}$ with respect to the corresponding conditions $y_i^{(j)}$ based on the dataset sizes $100,1\,000, 10\,000$ and $100\,000$. Specifically, it provides parity plots between target label values and true labels values obtained for generated $x^{(j),\mathrm{Gen}}$ for the label $Y_i=U_M$. Visually, CFM shows vast outperformance compared to INN, CWGAN and Bi approaches on datasets surpassing 1000 datapoints.
While INNs tend to produce fewer but more pronounced low-quality outliers compared to CWGANs, their average feature quality exceeds that of the CWGAN on all dataset sizes. In contrast, Bayesian inference exhibits the worst characteristics of both approaches, combining a high frequency of low accuracy generations with an overall broadly dispersed quality distribution.
Analogous effects are observed for the labels $\p$ and $G$, though to a lesser extent, see \autoref{app:parity_plots}.

On the labels $\p$ and $G$, accuracy of CWGAN, INN and BI models shows moderate differences on dataset sizes greater than $500$, while larger differences can be observed for $d=100$. On $Y_i=U_M$, the BI approach performs significantly worse than INN and CWGAN, both of which again provide similar accuracies.

\subsection{Diversity} \label{sec_val_diversity}
\begin{figure}
    \centering
    \includegraphics[width=\textwidth]{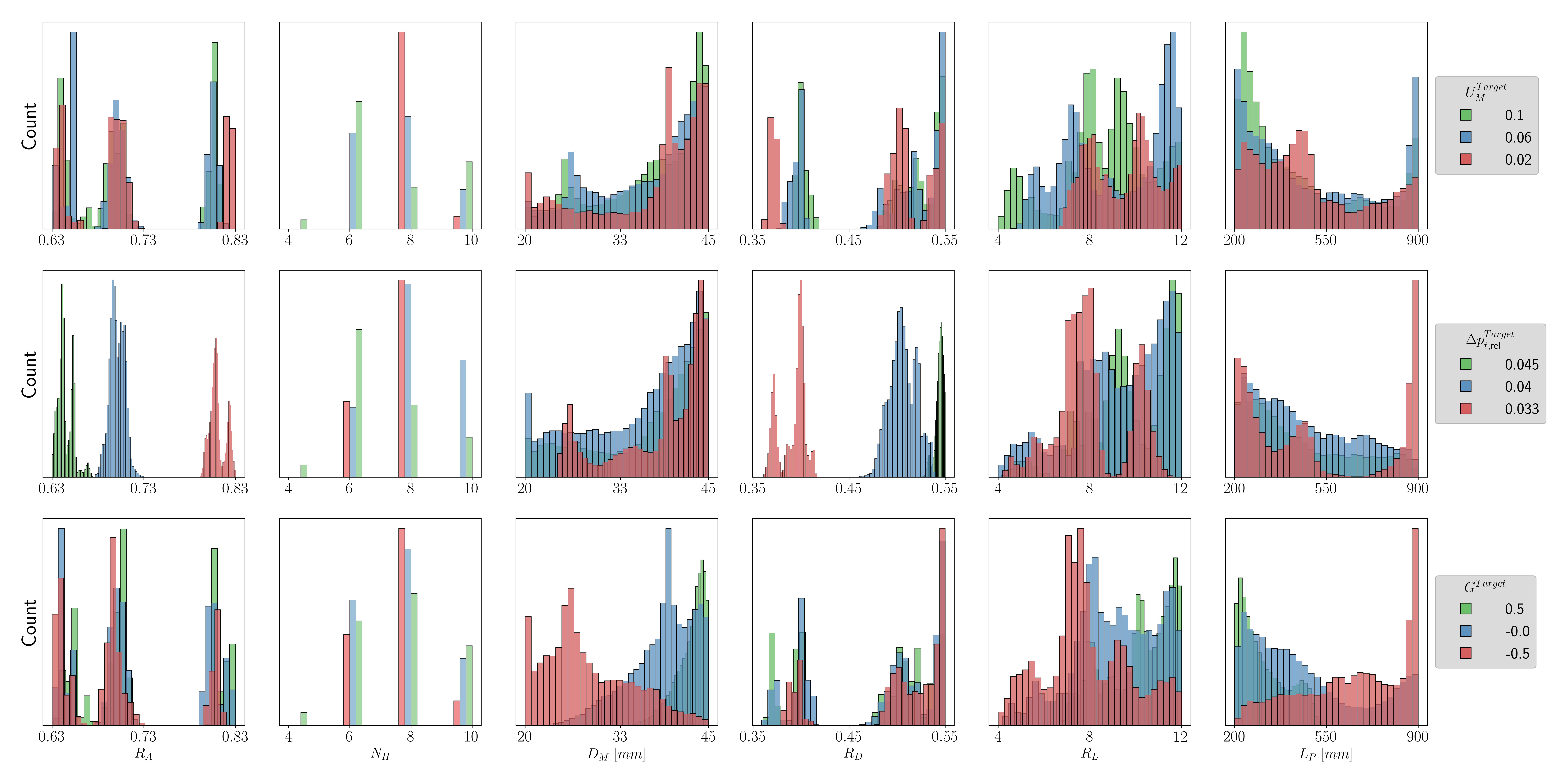}
    \caption{Distributions of the independent parameters $X^{\mathrm{Gen}}$ of generated by the CFM model conditioned target label values $Y_i^{\mathrm{Target}}$ and their respective values (top: $U_M$, center: $\p$, bottom: $G$).}
    \label{fig:param_dist_CFM}
\end{figure}

\begin{figure}
    \centering
    \includegraphics[width=\textwidth]{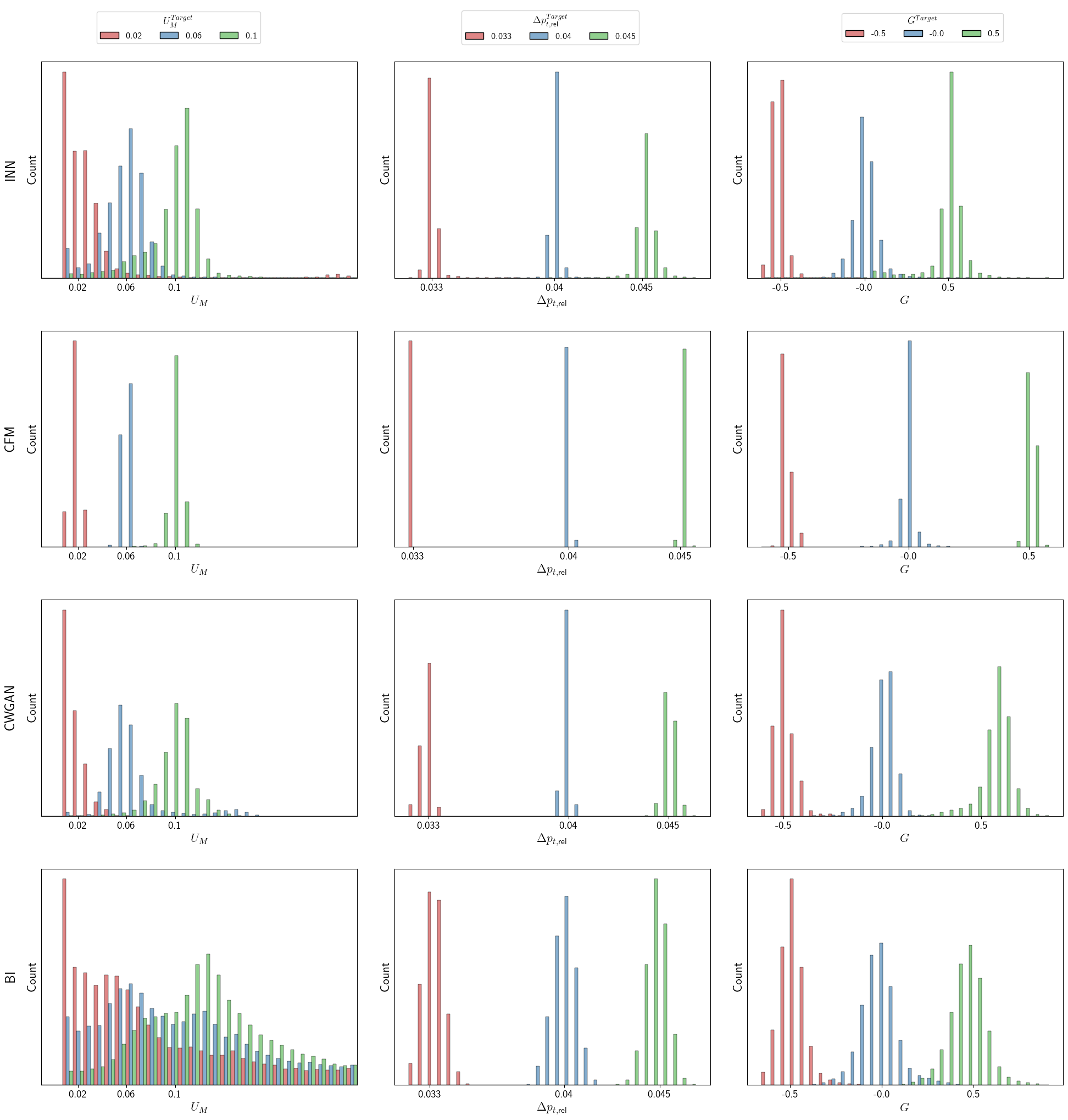}
    \caption{Distributions of true label values $Y_i^{\mathrm{Gen}}$ obtained for generated designs by all generative models (rows) for target labels $Y_i^{\mathrm{Target}}$ (left: $U_M$, center: $\p$, right: $G$), colored by target values. }
    \label{fig:label_dist_all_models}
\end{figure}

A second study was conducted to measure accuracy and diversity when generating a large number of designs for fixed targets. For each label $Y_i\in\{U_M,\p,G\}$, three target values $y_i^{\mathrm{Target}}$ were defined. For each of the 27 target vectors obtained from all combinations of target label values listed in \autoref{tab:diversity}, $5\,000$ designs $x^{\mathrm{Gen}}$ were generated by the respective generative models trained on $d=5\,000$ datapoints. True label values $Y^{\mathrm{Gen}}$ were again obtained using the surrogate models. For the CFM model, an overview of the distributions of generated design parameters depending on target values for each label (top: $U_M$, center: $\p$, bottom: $G$) is given by \autoref{fig:param_dist_CFM}. For orientation, the red histogram plots in the first row denote the parameter distributions of $X^{\mathrm{Gen}}$ generated for all $9$ target vectors with  $U_M^{\mathrm{Target}}=0.02$. 
The figure shows that geometrically diverse designs are obtained for a set choice of target labels. Different choices of targets influence the distributions of generated parameters in agreement with correlations observed in the dataset, see \autoref{fig:Figure_Dataset}. For instance, low target pressure loss $\p^\mathrm{Target}$ leads to designs with a high value of $R_A$ and low values of $R_D$. If $\p^{\mathrm{Target}}$ is high, these results are inverted. Similar clear relations between target label values and generated parameter distributions are visible for $N_H$ and $D_M$ which are strongly influenced by target values for $U_M$ and $G$, respectively. Similar correlations are observed for the other generative models (see \autoref{app:Diversity}). Nevertheless, the corresponding parameter distributions differ across models in terms of the sharpness and intensity of modes. This variability is further illustrated in \autoref{fig:label_dist_all_models}, which compares the parameter distributions generated by all models (rows) for three distinct target label vectors (colors).
In analogy to figure \autoref{fig:param_dist_CFM}, distributions of true values  $Y_i^{\mathrm{Gen}}$ are displayed for all labels (rows) and models (columns) in \autoref{fig:label_dist_all_models}, while a numerical evaluation by mean and standard deviation of true label values is given by table \autoref{tab:diversity}.

\begin{figure}[!t]
    \centering
    \includegraphics[width=\textwidth]{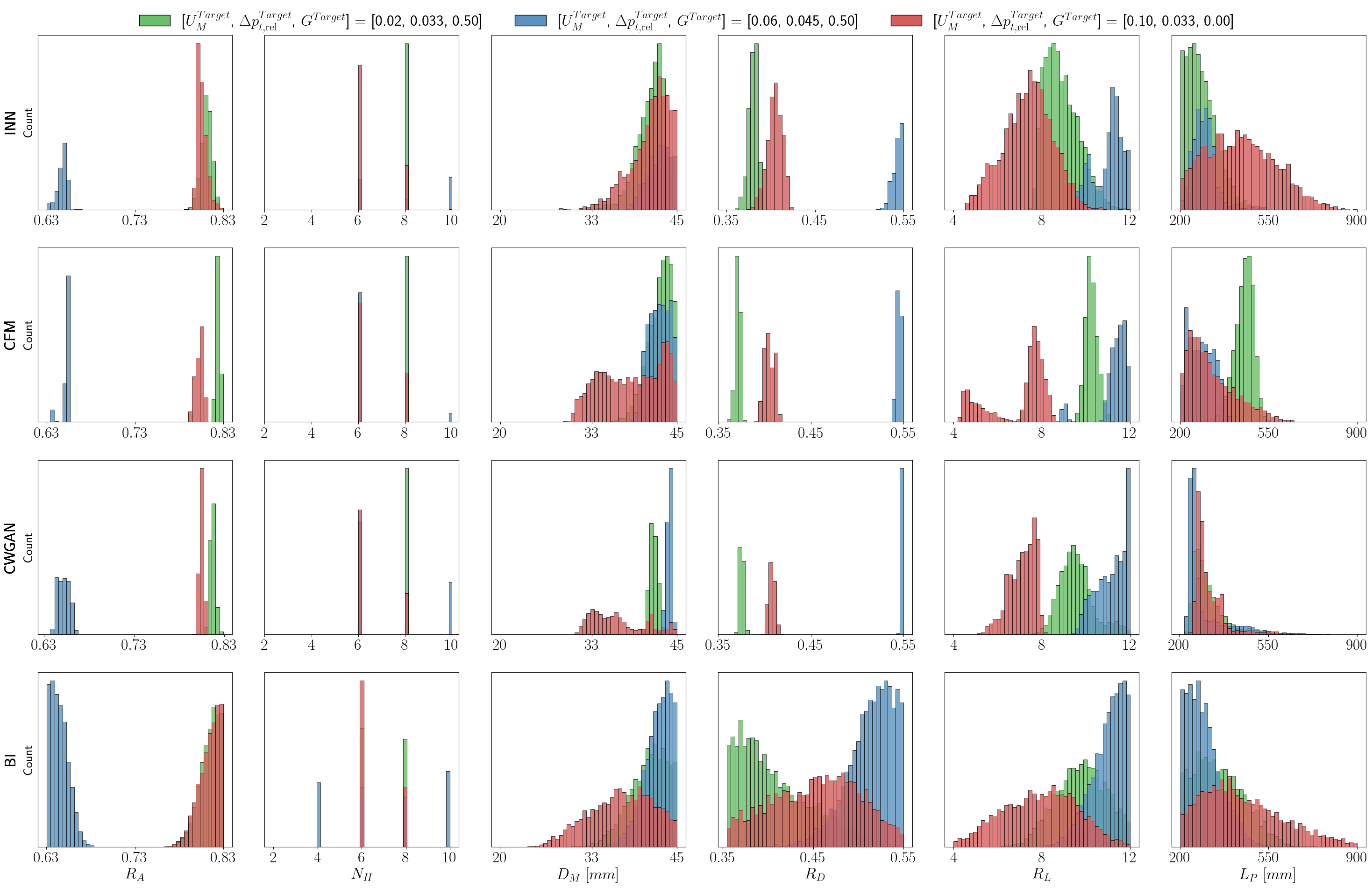}
    \caption{Distributions of generated parameters $X^\mathrm{Gen}$ by all generative models (rows) for three fixed target label vectors $Y^\mathrm{Target}$ (colors).}
\end{figure}

\begin{table}[!htb]
\centering
\begin{tabular}{c|c|cccccccc}
\toprule
                       & Model & \multicolumn{2}{c}{INN} & \multicolumn{2}{c}{CFM} & \multicolumn{2}{c}{CWGAN} & \multicolumn{2}{c}{BI} \\
\cline{3-10}
$y_i^\mathrm{Target}$ & Target
& $\mu$ & $\sigma$
& $\mu$ & $\sigma$
& $\mu$ & $\sigma$
& $\mu$ & $\sigma$ \\
\midrule
\multirow{3}{*}{$U_M$}
& 0.0200
& 0.0299 & 0.0353
& 0.0204 & 0.0037
& 0.0189 & 0.0073
& 0.0862 & 0.0730
 \\
& 0.0600
& 0.0584 & 0.0255
& 0.0597 & 0.0029
& 0.0628 & 0.0238
& 0.1112 & 0.0715
 \\
& 0.1000
& 0.0968 & 0.0262
& 0.0989 & 0.0055
& 0.0984 & 0.0147
& 0.1379 & 0.0666
 \\
\midrule
\multirow{3}{*}{$\p$}
& 0.0330
& 0.0332 & 0.0003
& 0.0330 & 0.0000
& 0.0330 & 0.0003
& 0.0333 & 0.0005
 \\
& 0.0400
& 0.0400 & 0.0003
& 0.0400 & 0.0001
& 0.0399 & 0.0002
& 0.0400 & 0.0006
 \\
& 0.0450
& 0.0451 & 0.0005
& 0.0450 & 0.0001
& 0.0449 & 0.0003
& 0.0447 & 0.0005
 \\
\midrule
\multirow{3}{*}{$G$}
& -0.5000
& -0.4960 & 0.0353
& -0.4932 & 0.0178
& -0.4798 & 0.0481
& -0.4742 & 0.0687
 \\
& 0.0000
& 0.0019 & 0.0758
& -0.0038 & 0.0310
& 0.0052 & 0.0580
& -0.0156 & 0.1004
 \\
& 0.5000
& 0.4818 & 0.1132
& 0.4963 & 0.0162
& 0.5645 & 0.0739
& 0.4523 & 0.0976
 \\
\bottomrule
\end{tabular}

\caption{Means and standard deviations of true label values for generated deigns $X^\mathrm{Gen}$ conditioned on given target label values.}
\label{tab:diversity}
\end{table}

In agreement with the first study conducted in \autoref{sec_val_accuracy}, CFM models yield the best performance. It can however be observed that, while the variance is noticeably higher, all other generative models still yield true label distributions centered around the respective target value on $\p$ and $G$. With $U_M$, the difference in accuracy between CFM and other models is most significant. While INN and CWGAN still yield results that are centered around the respective targets, albeit with more stark differences in variance compared to CFM, even more inaccuracy is observed with the Bayesian Inverse approach. Specifically for the label $U_M$, the true label distributions deteriorate substantially, showing weak alignment with their respective targets and a pronounced increase in spread that leads to substantial overlap with distributions associated with all other target values. This weak alignment and increased spread are also reflected in the corresponding mean and standard deviation values reported in \autoref{tab:diversity}.

\section{Conclusion}
\label{sec:conclusion}
In this work, we investigated the application of various state of the art generative machine learning approaches in the context of inverse problems. Generative models tested included both discrete and continuous flow-based invertible models represented by Invertible Neural Networks and Conditional Flow Matching, respectively, as well as Generative Adversarial Networks. Additionally, we designed and tested a Bayesian inverse design approach using feedforward neural networks. The inverse problem considered is given by a generic gas turbine combustor geometry that is defined by a reduced set of six independent design parameters and characterized by three performance labels. Based on an initial dataset obtained by a CFD simulation workflow, surrogate models were trained to emulate the forward process of deriving performance labels from given design vectors, thus providing a quick and reliable way of generating larger training datasets, as well as a consistent ground truth for the validation of generative models. We performed experiments comparing the performance of generative models on the inverse problem, i.e. the generation of design vectors that yield specified performance values. Two validation studies investigated the accuracy of inverse solutions obtained for a wide variety of targets, as well as the geometrical diversity of designs proposed given a fixed performance requirement. All generative models produce various design alternatives that largely agree with the respective specified targets. Between models, considerable differences in accuracy are observed. Most notably, Conditional Flow Matching yields significantly higher confidence of generated design proposals on all considered targets. These results motivate further investigation of the application of continuous flow-based models in the field of inverse problems. Next steps could include higher-dimensional, more complex design tasks, as well as applications to real world data.

\paragraph{Acknowledgements.} This work has been supported by the Federal Ministy of Economic Affairs and Energy (BMWI) through grant no. 03EE5186B and by Siemens Energy. We thank R.\ Chan, M.\ de Campos, C.\ Drygala, D. Rochau and B.\ Werdelmann and  for interesting discussions and the AG Turbo consortium for an interesting research environment. Hanno Gottschalk also acknowledges financial support by the Germen research council through SPP2403 "Carnot Batteries" project GO 833/8-1 "Inverse aerodynamic design of turbo components for Carnot batteries by means of physics
informed networks enhanced by generative learning".
 
\newpage
\appendix

\section{Parity Plots}\label{app:parity_plots}
We provide parity plots between target label values and obtained true labels for generated designs in analogy to \autoref{fig:parity_plot_comparison_u_m_limited_window} for the remaining labels $\p$ and $G$.

\begin{figure}[!htbp]
    \centering
    \includegraphics[width=0.9\textwidth]{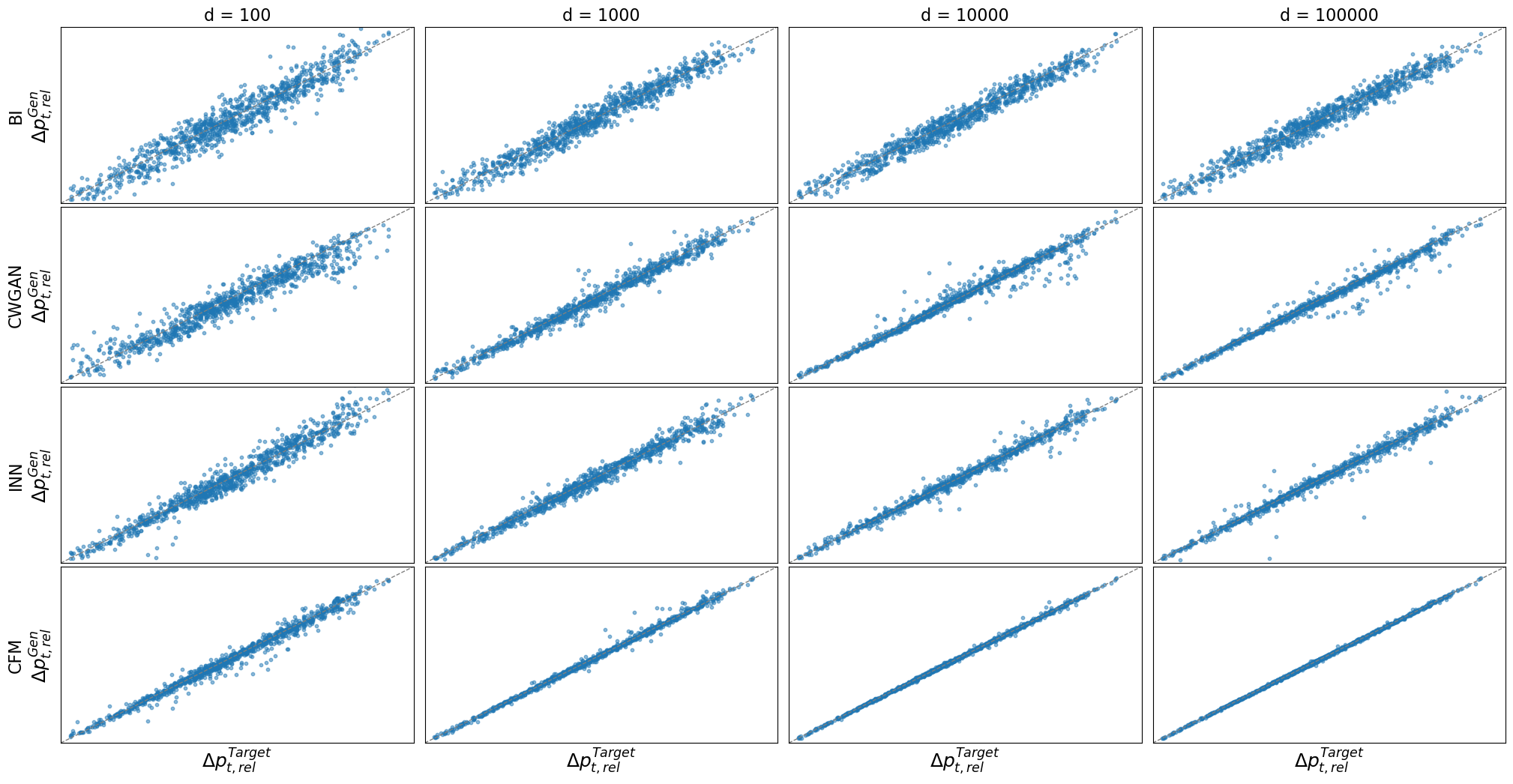}

    \vspace{0.5em}

    \includegraphics[width=0.9\textwidth]{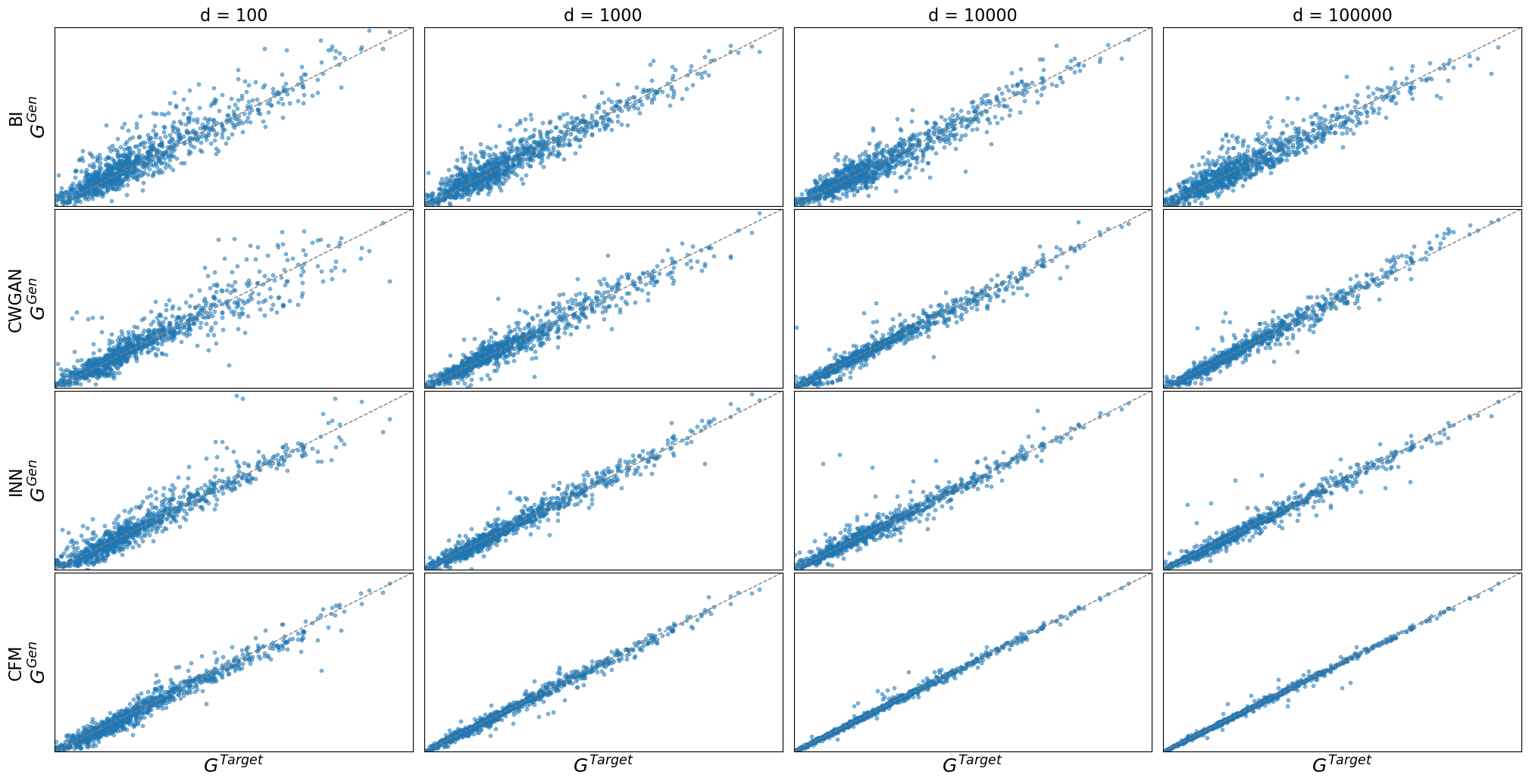}
    \caption{Parity plots between target and true label values for $Y_i=\p$ and $Y_i=G$ with $d\in\{100,1\,000, 10\,000,100\,000\}$.}
    \label{fig:parity_plots_combined}
\end{figure}

\FloatBarrier
\section{Distribution Plots}\label{app:Diversity}
We provide the distribution plots of generated parameters $X^\mathrm{Gen}$ for all target label vectors as in \autoref{fig:param_dist_CFM} for the remaining generative models.
\begin{figure}[!htbp]
    \centering
    \includegraphics[width=\textwidth]{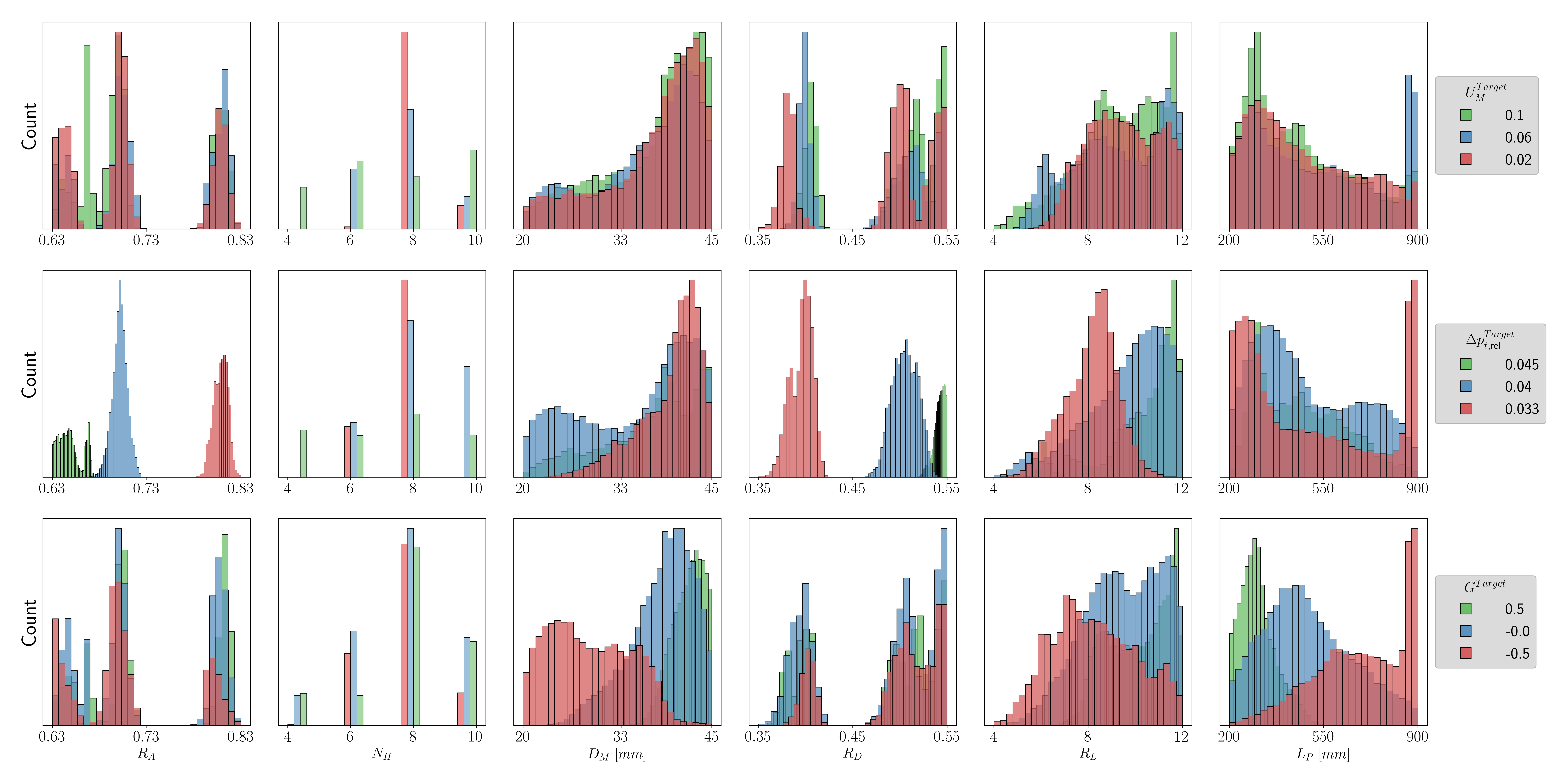}
    \caption{Distributions of the independent parameters $X^{\mathrm{Gen}}$ of generated by the INN model conditioned target label values $Y_i^{\mathrm{Target}}$ and their respective values (top: $U_M$, center: $\p$, bottom: $G$).}
    \label{fig:param_dist_INN}
\end{figure}

\begin{figure}[!htbp]
    \centering
    \includegraphics[width=\textwidth]{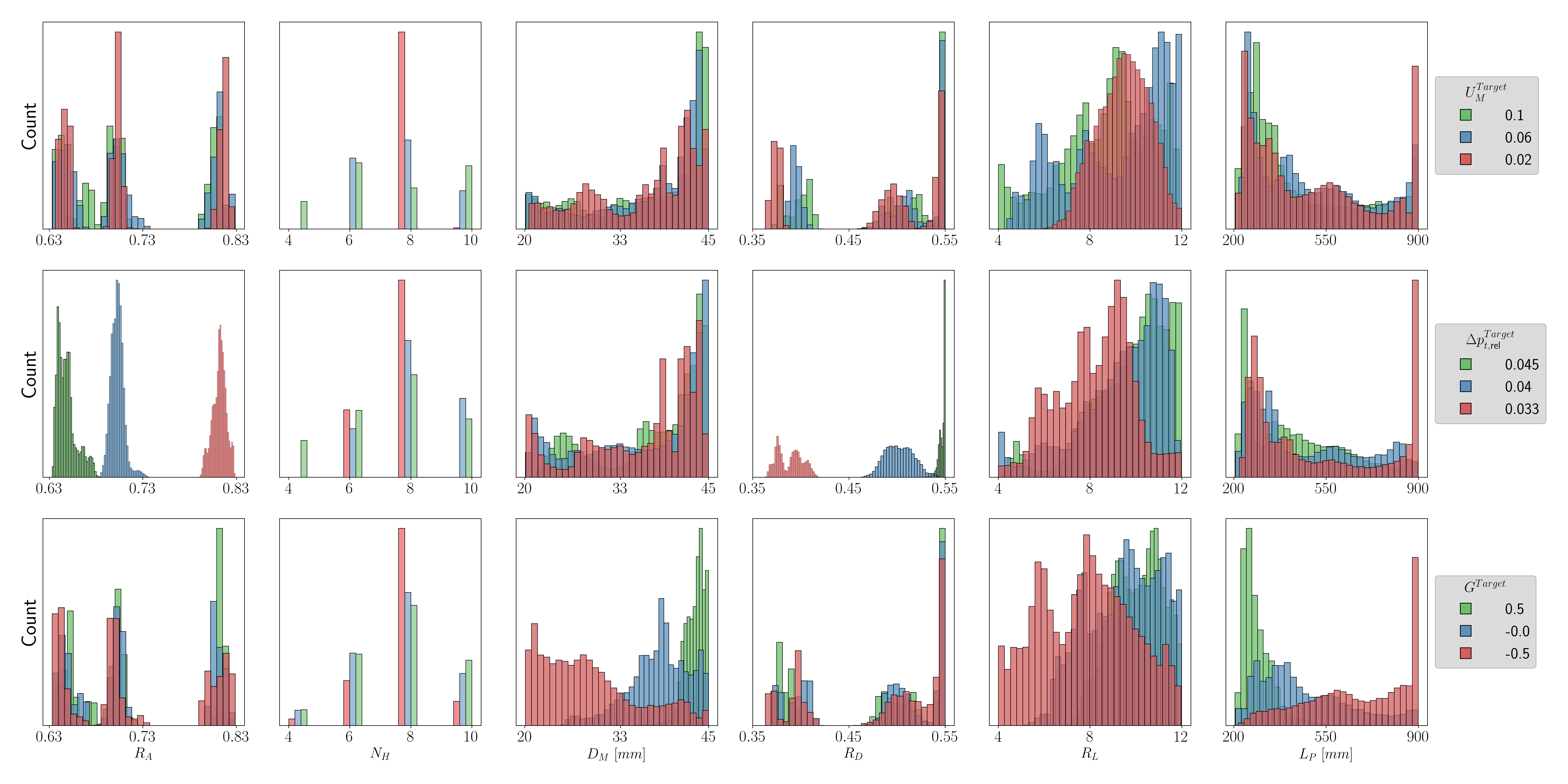}
    \caption{Distributions of the independent parameters $X^{\mathrm{Gen}}$ of generated by the CWGAN model conditioned target label values $Y_i^{\mathrm{Target}}$ and their respective values (top: $U_M$, center: $\p$, bottom: $G$).}
    \label{fig:param_dist_WGAN}
\end{figure}

\begin{figure}[!htbp]
    \centering
    \includegraphics[width=\textwidth]{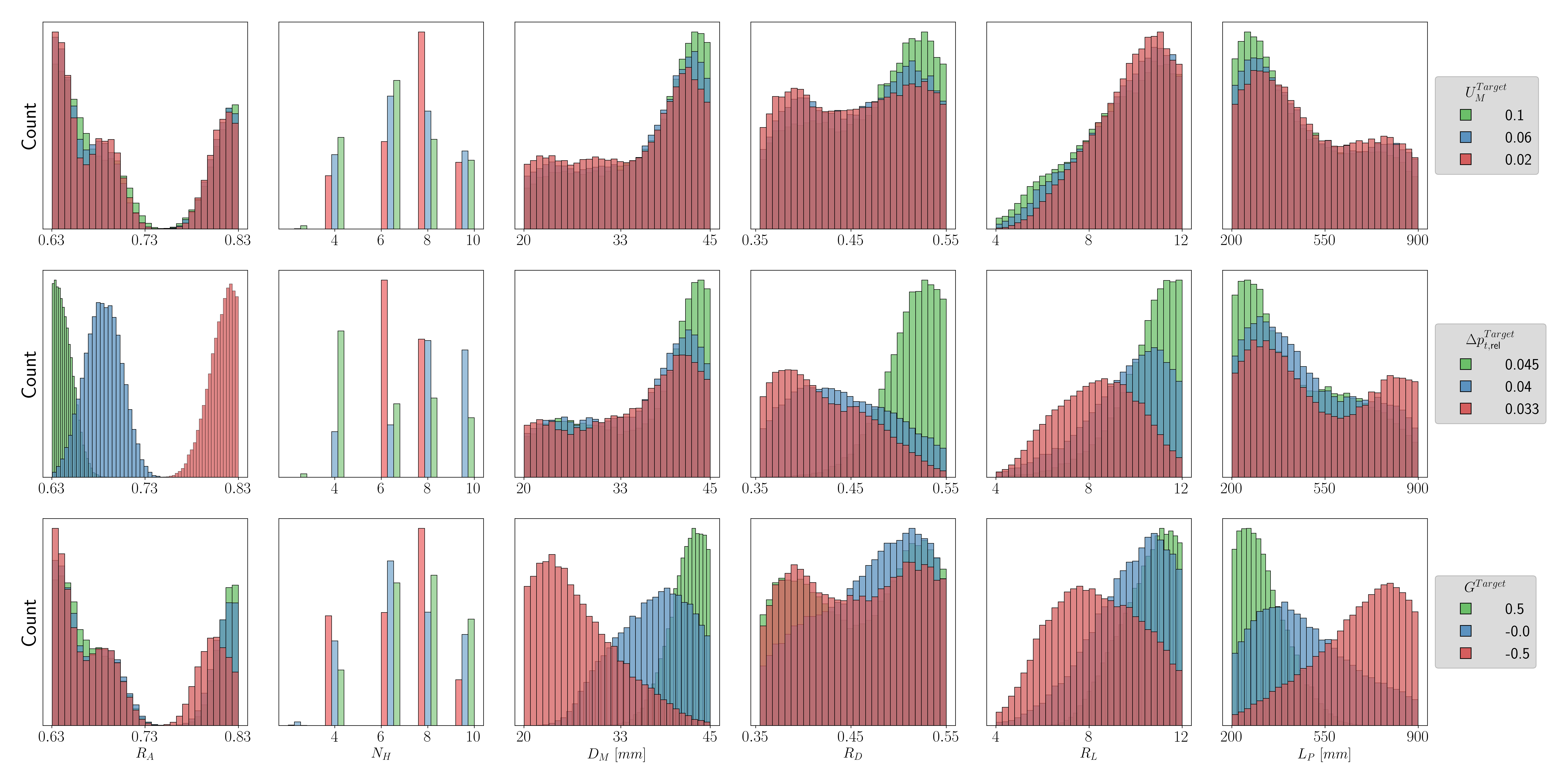}
    \caption{Distributions of the independent parameters $X^{\mathrm{Gen}}$ of generated by the BI model conditioned target label values $Y_i^{\mathrm{Target}}$ and their respective values (top: $U_M$, center: $\p$, bottom: $G$).}
    \label{fig:param_dist_MCMC}
\end{figure}

\FloatBarrier  

\bibliography{sample}
\end{document}